\documentclass{article}

\usepackage[preprint,nonatbib]{neurips_2022}

\usepackage[utf8]{inputenc} 
\usepackage[T1]{fontenc}    
\usepackage{hyperref}       
\usepackage{url}            
\usepackage{booktabs}       
\usepackage{amsfonts}       
\usepackage{nicefrac}       
\usepackage{microtype}      
\usepackage{xcolor}         
\usepackage{amsmath}
\usepackage{amsfonts}
\usepackage{graphicx}
\usepackage{algorithm}
\usepackage{algorithmic}

\title{Clipped-Objective Policy Gradients for Pessimistic Policy Optimization}

\author{%
  Jared Markowitz \\
  Johns Hopkins University\\
  Applied Physics Laboratory\\
  Laurel, MD 20723\\
  \texttt{Jared.Markowitz@jhuapl.edu} \\
  \And
  Edward W. Staley \\
  Johns Hopkins University\\
  Applied Physics Laboratory\\
  Laurel, MD 20723\\
  \texttt{Edward.Staley@jhuapl.edu} \\
}

\begin{document}

\maketitle

\begin{abstract}
  To facilitate efficient learning, policy gradient approaches to deep reinforcement learning (RL) are typically paired with variance reduction measures and strategies for making large but safe policy changes based on a batch of experiences. Natural policy gradient methods, including Trust Region Policy Optimization (TRPO), seek to produce monotonic improvement through bounded changes in policy outputs. Proximal Policy Optimization (PPO) is a commonly used, first-order algorithm that instead uses loss clipping to take multiple safe optimization steps per batch of data, replacing the bound on the single step of TRPO with regularization on multiple steps. In this work, we find that the performance of PPO, when applied to continuous action spaces, may be consistently improved through a simple change in objective. Instead of the importance sampling objective of PPO, we instead recommend a basic policy gradient, clipped in an equivalent fashion. While both objectives produce biased gradient estimates with respect to the RL objective, they also both display significantly reduced variance compared to the unbiased off-policy policy gradient. Additionally, we show that (1) the clipped-objective policy gradient (COPG) objective is on average ``pessimistic'' compared to both the PPO objective and (2) this pessimism promotes enhanced exploration. As a result, we empirically observe that COPG produces improved learning compared to PPO in single-task, constrained, and multi-task learning, without adding significant computational cost or complexity. Compared to TRPO, the COPG approach is seen to offer comparable or superior performance, while retaining the simplicity of a first-order method.
\end{abstract}

\section{Introduction}
Several strategies are available for model-free deep reinforcement learning (DRL).  Some optimize a policy directly \cite{Wi92, schulman2017trust, a3c,
 ScWoDhRaKl17}, computing an estimate of the gradient of expected reward with respect to the parameters of the policy and using it to perform stochastic gradient descent.  Others use neural networks to approximate state-action ($Q$) values, which are then used to formulate a policy (e.g., \cite{qnature, ddpg, td3, sac})  Both approaches have benefits and drawbacks. Value-based methods are typically formulated off-policy, enabling reuse of data collected throughout training and providing superior sample efficiency.  However, value-based methods that leverage function approximation lack theoretical guarantees and are typically less easily parallelized than on-policy methods centered on policy gradient estimates.

To improve learning efficiency, leading policy gradient algorithms include measures to (1) reduce the variance of the policy gradient estimate and (2) take one or more optimization steps that are large but safe, for each batch of collected data.  Strategies for reducing variance include the use of learned value networks as baselines and generalized advantage estimation \cite{ScMoLeJoAb16}.  Relatively large but safe optimization steps may be achieved through natural policy gradient / trust region methods, which seek monotonic improvement through approximate policy iteration and a constraint on how much the policy network outputs may change \cite{kakade_npg, natural, schulman2017trust}. 

Alternatively, multiple small optimization steps may be taken on a single batch, using regularization and sometimes a stopping condition \cite{ScWoDhRaKl17}.  Proximal Policy Optimization (PPO; \cite{ScWoDhRaKl17}) is a leading approach that employs this strategy, relying on clipping-based regularization to take multiple optimization steps per batch of sampled data, when appropriate.  PPO was formulated as a first-order alternative to Trust Region Optimization (TRPO; \cite{schulman2017trust}), which falls in the former category (seeking monotonic improvement through constrained, approximate policy iteration). While less theoretically grounded than TRPO, PPO is simpler to implement, allows for sharing of policy and value network layers, and typically provides similar performance.  Consequently, PPO has become a frequently-used, general-purpose approach.

Here we propose a small modification to PPO that we find consistently improves its performance in problems with continuous action spaces.  Instead of the clipped ``surrogate'' objective used in PPO, we recommend a basic policy gradient objective clipped in a similar manner.  The clipping mechanism of PPO is designed to be ``pessimistic," favoring faster migration away from problematic parts of state space over migration toward beneficial parts.  We show that our approach furthers this tendency, leading to further reduction in the likelihood of premature convergence to suboptimal solutions and consistently improved outcomes.  
 
In the following, we first motivate and specify our alternative objective.  We then demonstrate the benefits it confers in standard, constrained, and multi-task settings.

\section{Related Work}
Policy gradient methods were initially popularized by REINFORCE \cite{Wi92}, and have been significantly improved in recent years. Updated methods \cite{schulman2017trust, a3c, ScWoDhRaKl17} reduce variance on the policy gradient estimate through the use of reward-to-go, a state-dependent critic, and advantage estimation.  Natural policy gradient methods \cite{kakade_npg, natural} improve upon the basic approach, considering policy changes in terms of differences in network output rather than differences in network parameters.  This allows the large plateaus that frequently occur in return space to be handled more effectively, and is integrated into Trust Region Policy Optimization \cite{schulman2017trust}.  For each batch, TRPO uses the natural policy gradient to improve the policy subject to a constraint in KL divergence, via backtracking line search.  Proximal Policy Optimization \cite{ScWoDhRaKl17} replaces this constrained optimization with pessimistic clipping-based regularization and multiple gradient steps per batch of data.  It may be paired with an early-stopping condition based on a limit in the average KL divergence between the current policy and the policy that generated the data (e.g., in \cite{SpinningUp2018}).

\section{Preliminaries: Policy Gradients and PPO}

Below we briefly recap policy gradient methods, with the goal of motivating our alternative approach to safely taking multiple optimization steps per batch of collected data.

\subsection{Basic Policy Gradient}
To learn an optimal policy, the reinforcement learning objective
\begin{equation}
J(\theta) =  \mathbb{E}_{\tau \sim p_\theta(\tau)} r(\tau)
\label{rl}
\end{equation}
may be directly differentiated \cite{Wi92}.
Here $p_\theta(\tau)$ is the probability distribution over trajectories $\tau \equiv \mathbf{s}_1, \mathbf{a}_1, \ldots,  \mathbf{s}_T, \mathbf{a}_T $ induced by a policy parameterized by $\theta$;  $r(\tau) \equiv \sum_t r(\mathbf{s}_t, \mathbf{a}_t) $ is the total reward of a given trajectory with states $\mathbf{s}_t$, actions $\mathbf{a}_t$, and rewards $r(\mathbf{s}_t, \mathbf{a}_t)$. The gradient of this quantity with respect to policy parameters $\theta$  may be evaluated by writing out the joint probability and expectation explicitly, using the expression for the derivative of the logarithm, and leveraging the independence of the probability distribution of initial state and the state transition functions on $\theta$: 
\begin{equation}
\begin{split}
\nabla_\theta J(\theta) &= \nabla_\theta \int p_\theta(\tau) r(\tau) d\tau = \int p_\theta(\tau)\nabla_\theta \log p_\theta(\tau) r(\tau) d\tau\\
& = \mathbb{E}_{\tau \sim p_\theta(\tau)} \nabla_\theta \left[ \log p_\theta(\tau) \right]r(\tau) = \mathbb{E}_{\tau \sim p_\theta(\tau)} \nabla_\theta \left[ \log \pi_\theta(\tau) \right]r(\tau)  .
\end{split}
\end{equation}
Here $\pi_\theta(\tau)$ refers to the product of each individual policy evaluation $\pi_\theta(\mathbf{a}_t | \mathbf{s}_t)$ in the trajectory.  Both the initial distribution of states $p(\mathbf{s}_1)$ and the state transition probabilities $p(\mathbf{s}_{t+1} | \mathbf{s}_t, \mathbf{a}_t)$ are independent of $\theta$.  This expression may be sampled, leading to the basic policy gradient used in REINFORCE \cite{Wi92}:
\begin{equation}
\nabla_\theta J(\theta) \approx \frac{1}{N} \sum_{i=1}^N  \Bigg[ \bigg(\sum_{t=1}^{T_i}\nabla_\theta \log \pi_\theta(\mathbf{a}_{i,t}| \mathbf{s}_{i,t}) \bigg)\bigg(\sum_{t = 1}^{T_i}r( \mathbf{s}_{i,t}, \mathbf{a}_{i,t})\bigg)\Bigg]
\end{equation}
This expression provides an unbiased but high-variance estimate of the policy gradient for expected reward.  Variance may be reduced without introducing bias through the use of reward-to-go and a state-dependent baseline.   The form of policy gradient most commonly considered is of the form
\begin{equation}
\nabla_\theta J(\theta) \approx \frac{1}{N} \sum_{i=1}^N  \sum_{t=1}^{T_i}\nabla_\theta \log \pi_\theta(\mathbf{a}_{i,t}| \mathbf{s}_{i,t}) \hat{A}^\pi(\mathbf{s}_{i,t}, \mathbf{a}_{i,t}), \label{eq:pg}
\end{equation}
where $\hat{A}^\pi(\mathbf{s}_{i,t}, \mathbf{a}_{i,t})$ is an estimate of the advantage of a given action.  Different choices are available for the form of the advantage function. Typically discount factors and some bootstrapping are incorporated, which introduce bias but substantially decrease variance.  As is done in PPO, we use Generalized Advantage Estimation (GAE; \cite{ScMoLeJoAb16}), which enables explicit tuning of the bias-variance trade-off of the advantage estimate.


\subsection{Off-Policy Policy Gradient}
Despite gains made through variance reduction, policy gradient methods remain data-intensive.  A leading contributor to this deficiency is their inability to learn from data collected under any but the current policy.  In contrast to this \emph{on-policy} approach, \emph{off-policy} methods may be updated based on data collected using other policies.  This can lead to significant efficiency gains, through the reuse of data.

An off-policy version of the policy gradient may be derived through the following \cite{Precup2000, Precup2001, levine_lecture}.  Importance sampling may be used to write the reinforcement learning objective evaluated for policy parameters $\theta'$ using data collected with policy parameters $\theta$ as
\begin{equation}
J(\theta') = \mathbb{E}_{\tau \sim p_\theta(\tau)} \bigg[ \frac{p_{\theta'}(\tau)}{p_{\theta}(\tau)}r(\tau)\bigg].
\end{equation}
Here the probabilities $p_{\theta'}(\tau)$ and $p_{\theta}(\tau)$ are the probabilities an agent will traverse trajectory $\tau$ given policy parameters $\theta'$ or $\theta$, respectively. The policy gradient of this quantity is
\begin{equation}
    \begin{split}
    \nabla_{\theta'}J(\theta') &= \mathbb{E}_{\tau \sim p_\theta(\tau)} \Bigg[ \frac{p_{\theta'}(\tau)}{p_{\theta}(\tau)} \nabla_{\theta'}\log \pi_{\theta'}(\tau)r(\tau)\Bigg]\\
    &= \mathbb{E}_{\tau \sim p_\theta(\tau)} \Bigg[ \bigg( \prod_{t=1}^T \frac{\pi_{\theta'}(\mathbf{a}_t | \mathbf{s}_t)}{\pi_{\theta}(\mathbf{a}_t | \mathbf{s}_t)}\bigg) \bigg( \sum_{t=1}^T\nabla_{\theta'}\log\pi_{\theta'}(\mathbf{a}_t|\mathbf{s}_t)\bigg)\bigg(\sum_{t=1}^Tr(\mathbf{s}_t, \mathbf{a}_t)\bigg)\Bigg].
    \end{split}
\end{equation}
Accounting for causality, 
\begin{equation}
    \begin{split}
    \nabla_{\theta'}J(\theta') = \mathbb{E}_{\tau \sim p_\theta(\tau)} \Bigg[ \sum_{t=1}^T\nabla_{\theta'}\log\pi_{\theta'}(\mathbf{a}_t|\mathbf{s}_t) &\Bigg( \prod_{t'=1}^t \frac{\pi_{\theta'}(\mathbf{a}_{t'} | \mathbf{s}_{t'})}{\pi_{\theta}(\mathbf{a}_{t'} | \mathbf{s}_{t'})}\Bigg) * \\
    &\Bigg( \sum_{t'=t}^Tr(\mathbf{s}_{t'}, \mathbf{a}_{t'}) \Bigg( \prod_{t''=t}^{t'} \frac{\pi_{\theta'}(\mathbf{a}_{t''} | \mathbf{s}_{t''})}{\pi_{\theta}(\mathbf{a}_{t''} | \mathbf{s}_{t''})}\Bigg) \Bigg) \Bigg]. 
    \end{split}
    \label{oppg}
\end{equation}
Here the first product accounts for the differences in probability for reaching a given state / decision when acting under $\pi_\theta$ or $\pi_{\theta'}$, while the latter reflects the differences in probabilities for rewards being achieved from future transitions when acting under the two policies.  Unfortunately this form is prone to high variance, due to the products of fractions composed of numbers less than 1.   Further approximation is therefore required to achieve practical optimization.

\subsection{Proximal Policy Optimization}
As mentioned above, Proximal Policy Optimization \cite{ScWoDhRaKl17} was proposed as a first-order alternative to Trust Region Policy Optimization \cite{schulman2017trust}.  TRPO seeks monotonic performance improvement through approximate policy iteration, comparing the RL objective under the old and new policies via importance sampling, in a region specified by a constraint on the average KL divergence between the old and new policies over the current batch.  PPO shares the resulting importance sampling objective, but omits the constraint on KL divergence.  Instead, it allows for multiple small updates per batch and incorporates ``pessimistic'' clipping of the objective:
\begin{equation}
    J_{\text{PPO}}(\theta') = \mathbb{E}_t \bigg[ \min \bigg( \frac{\pi_{\theta'}(\mathbf{a}_t | \mathbf{s}_t)}{\pi_\theta(\mathbf{a}_t | \mathbf{s}_t)}\hat{A}^\pi(\mathbf{s}_t, \mathbf{a}_t), \text{clip}\bigg(\frac{\pi_{\theta'}(\mathbf{a}_t | \mathbf{s}_t)}{\pi_\theta(\mathbf{a}_t | \mathbf{s}_t)}, 1 \pm \epsilon\bigg) \hat{A}^\pi(\mathbf{s}_t, \mathbf{a}_t)\bigg)\bigg]
    \label{eq: ppo}
\end{equation}
The clipping factor $\epsilon$ is typically taken to be around $0.2$, and is sometimes decreased as training progresses.  This objective is referred to as ``pessimistic'' because it is a lower bound of the unclipped objective, preventing excessively large updates in the case of positive advantages and increasing the magnitude of small updates with negative advantages.  While derived differently, this objective represents an approximation of \ref{oppg}, wherein high-variance terms are removed \cite{levine_lecture}.  As mentioned above, PPO may also use early stopping \cite{SpinningUp2018} to approximate the KL divergence constraint of TRPO.

\section{Clipped-Objective Policy Gradient}
As an alternative to both the off-policy policy gradient and PPO, we propose the clipped-objective policy gradient (COPG).  Instead of clipping the policy ratio of the importance sampling objective, as is done by PPO, we instead clip the policy gradient objective \ref{eq:pg} in a similar fashion:
\begin{equation}
\begin{split}
J_\text{COPG}(\theta') = \mathbb{E}_t\bigg[\min\bigg(&\log \pi_{\theta'}(\mathbf{a}_{t}| \mathbf{s}_{t})\hat{A}^\pi(\mathbf{s}_{t}, \mathbf{a}_{t}),\\ 
&\log\bigg(\text{clip} \bigg(\frac{\pi_{\theta'}(\mathbf{a}_{t}| \mathbf{s}_{t})}{\pi_\theta(\mathbf{a}_{t}| \mathbf{s}_{t})}, 1 \pm \epsilon\bigg)\pi_\theta(\mathbf{a}_{t}| \mathbf{s}_{t})\bigg)\hat{A}^\pi(\mathbf{s}_{t}, \mathbf{a}_{t})\bigg)\bigg].
\label{clip}
\end{split}
\end{equation}
We then choose $\epsilon$ as one would in PPO and perform multiple updates per batch in a similar fashion.  

Beyond the first step (and putting aside the clipping), this objective effectively ignores the mismatch between the policy that generated the advantage estimates ($\pi_\theta$) and the policy that is being optimized ($\pi_{\theta'})$.  As in PPO the clipping behaves pessimistically, with the unclipped objective providing an upper bound on the clipped objective.  Early stopping based on KL divergence may also be used in conjunction with this objective.

Comparing the gradient of this objective with that of the PPO objective \ref{eq: ppo}, we see that 

\begin{equation}
    \frac{\nabla J_{\text{COPG}}}{ \nabla J_{\text{PPO}}}=\begin{cases}
          \pi_\theta / \pi_{\theta'} \quad &\text{if no clipping} \\
          1 / (1 \pm \epsilon) \quad &\text{if clipping}.
     \end{cases}\label{comp}
\end{equation}

The two updates are identical for the first update of each batch (when $\pi_{\theta'}=\pi_\theta$), but differ thereafter.  Unlike the off-policy policy gradient \ref{oppg}, both PPO and COPG provide biased updates toward the true RL objective.  However these biased updates have significantly reduced variance compared to the off-policy policy gradient, because they do not include quotients of policy terms.

\subsection{Why should this help?}
To see the potential benefits of the COPG objective, consider the ratios \ref{comp} and the migration of the policy terms $\pi_{\theta'}$ as the network is updated.  The policy gradient may be viewed as a weighted form of maximum likelihood, where the policy is driven toward regions of positive advantage and away from regions of negative advantage \cite{levine_lecture}.  On average then, $\pi_{\theta'}(\mathbf{a_t} | \mathbf{s_t}) > \pi_\theta(\mathbf{a_t} | \mathbf{s_t})$ for decisions with $\hat{A}^\pi > 0$, and vice versa.  Clipping will more frequently be on the high end ($1+\epsilon$) for  $\hat{A}^\pi > 0$, and vice versa.  Combining these observations with Equation \ref{comp}, we see that compared to PPO the COPG should, on average, take smaller steps toward decisions with positive advantage and larger steps away from decisions with negative advantage.  We might expect this form of ``pessimism'' to aid in agent exploration, helping to prevent premature convergence to a suboptimal solution.  If significant, this change should be observable in the form of typically higher policy entropy for the COPG than PPO throughout training.


\subsection{Algorithmic Formulation}
The clipped-objective policy gradient may be substituted into any formulation of PPO and used in exactly the same way as the standard objective, without significantly increasing implementation complexity or compute requirements.  Other modifications of the basic PPO algorithm (early stopping, minibatches, additional clipping, etc.) may likewise be used, and hyperparameters should not need to be changed.  One less-frequently used modification, the ``clipped action policy gradient'' correction of \cite{pmlr-v80-fujita18a}, may be particularly useful in some environments.  This correction properly accounts for the control bounds in continuous action spaces, and is applicable to any method that uses a multivariate normal distribution to model its policy.

\section{Experiments}
To evaluate the impact of the clipped-objective policy gradient, we compared its performance to that of PPO and TRPO in a variety of settings.  In each set of experiments COPG was configured identically to PPO, with only the objective function differing.  All hyperparameters were chosen in accordance with values previously established for PPO.  Five random seeds were trained for each single-task agent, while three random seeds were trained for each multi-task agent.

While this approach also applies to discrete action spaces, we chose to focus on multidimensional, continuous action spaces.  This choice was based on the availability of test environments and the ability to more naturally evaluate policy changes in continuous action spaces.

\subsection{Single-Task Learning}
\begin{figure}
\centering
\includegraphics[width=0.24\textwidth]{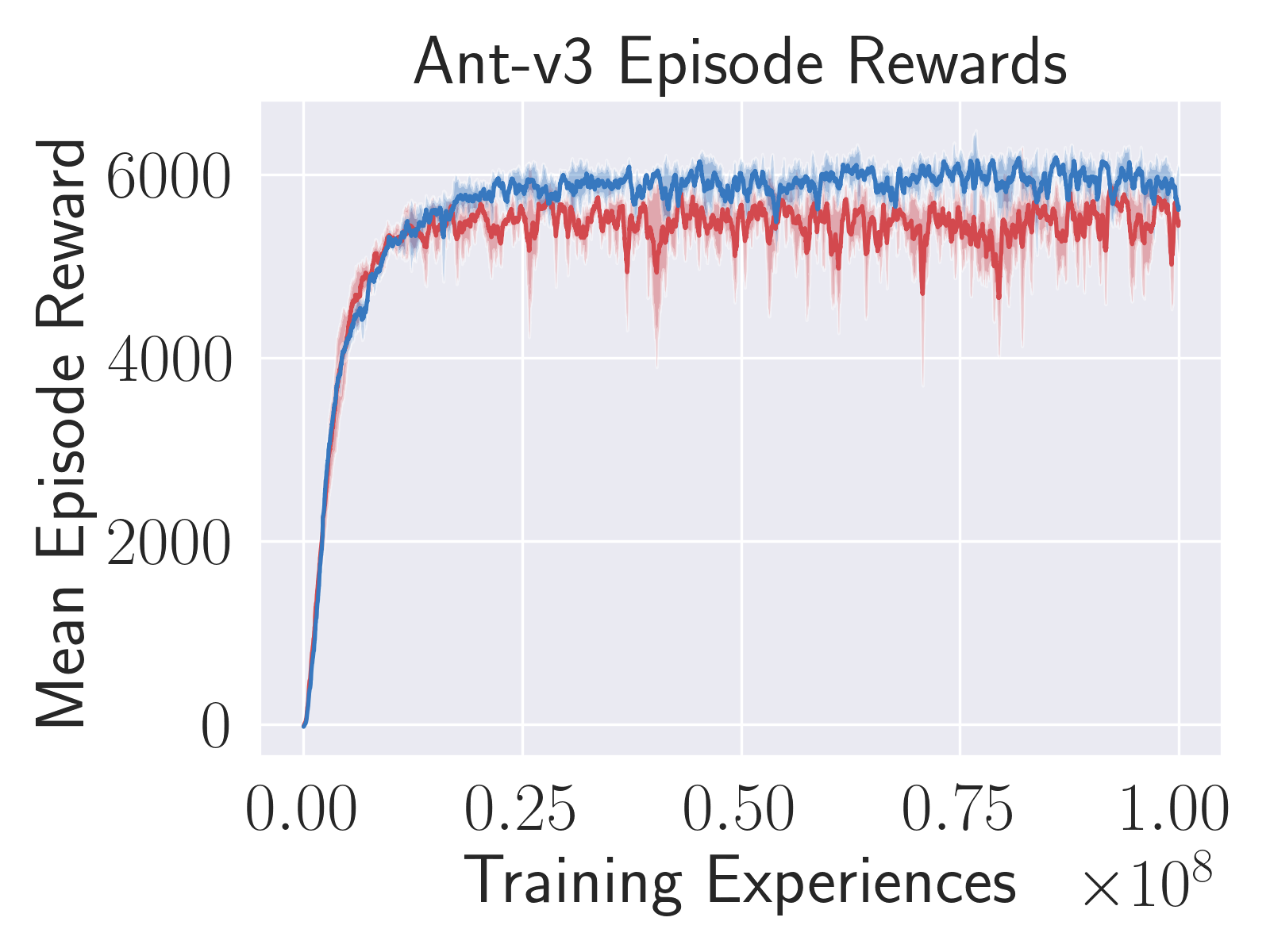}
\includegraphics[width=0.24\textwidth]{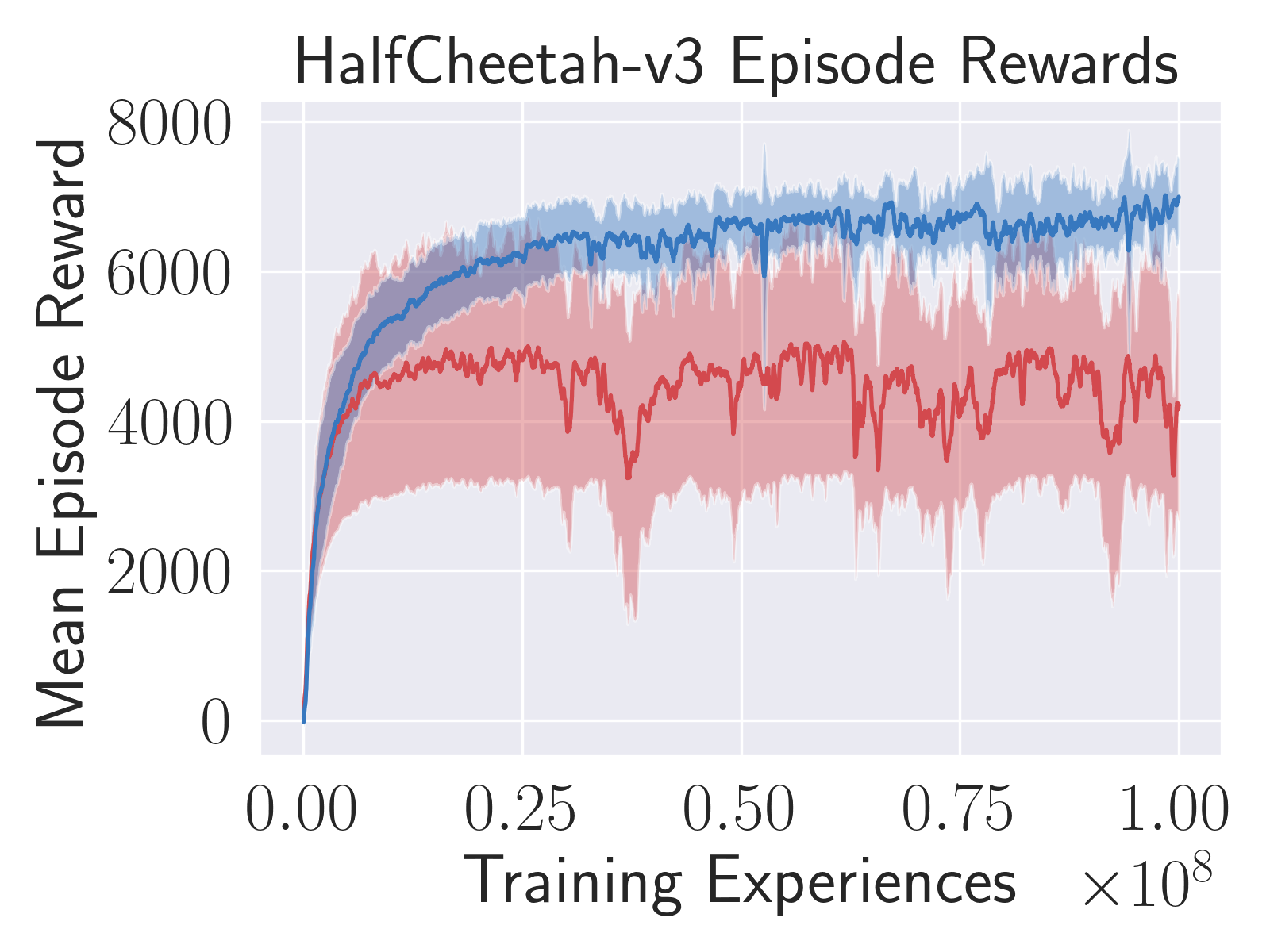}
\includegraphics[width=0.24\textwidth]{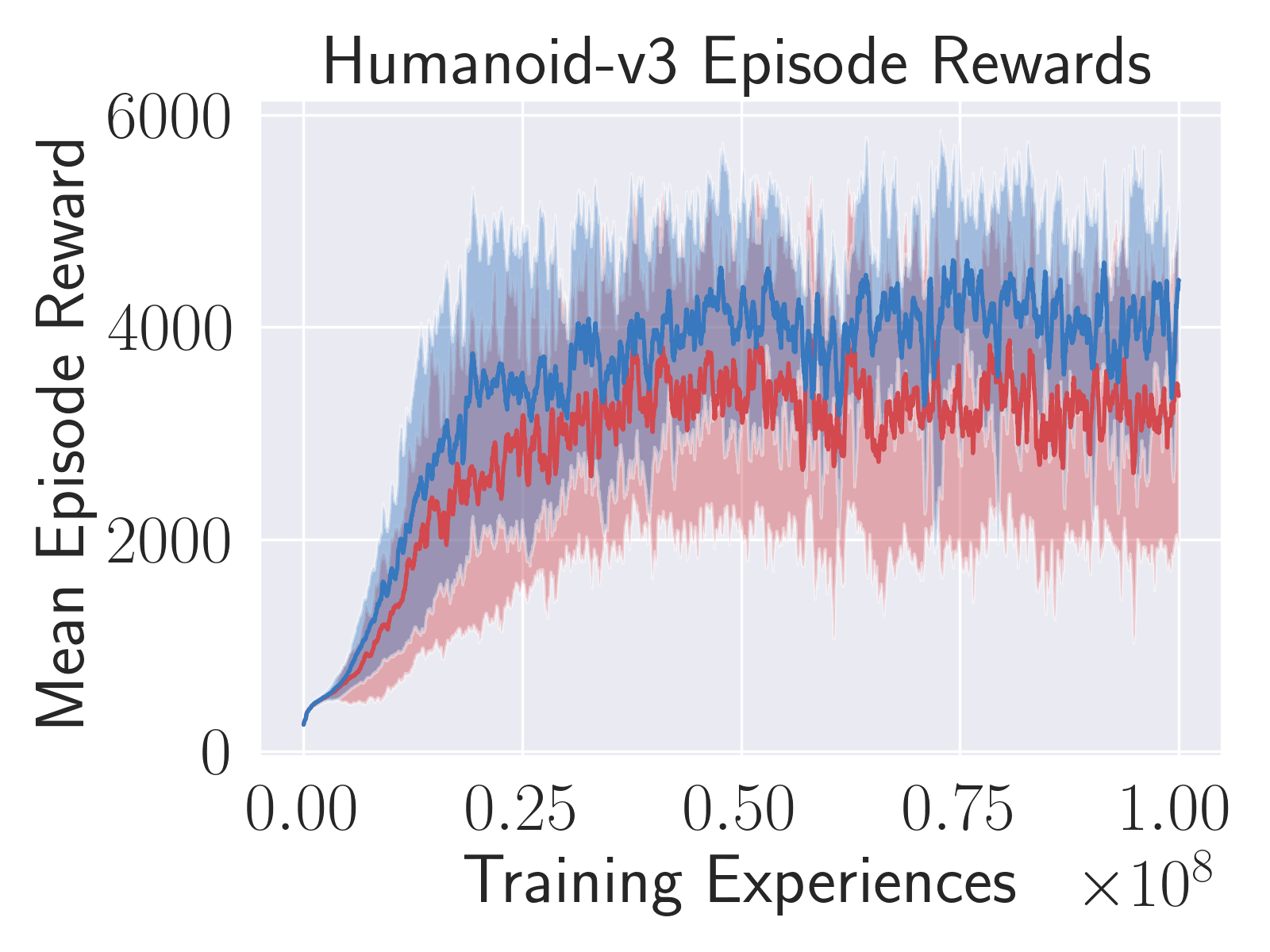}
\includegraphics[width=0.24\textwidth]{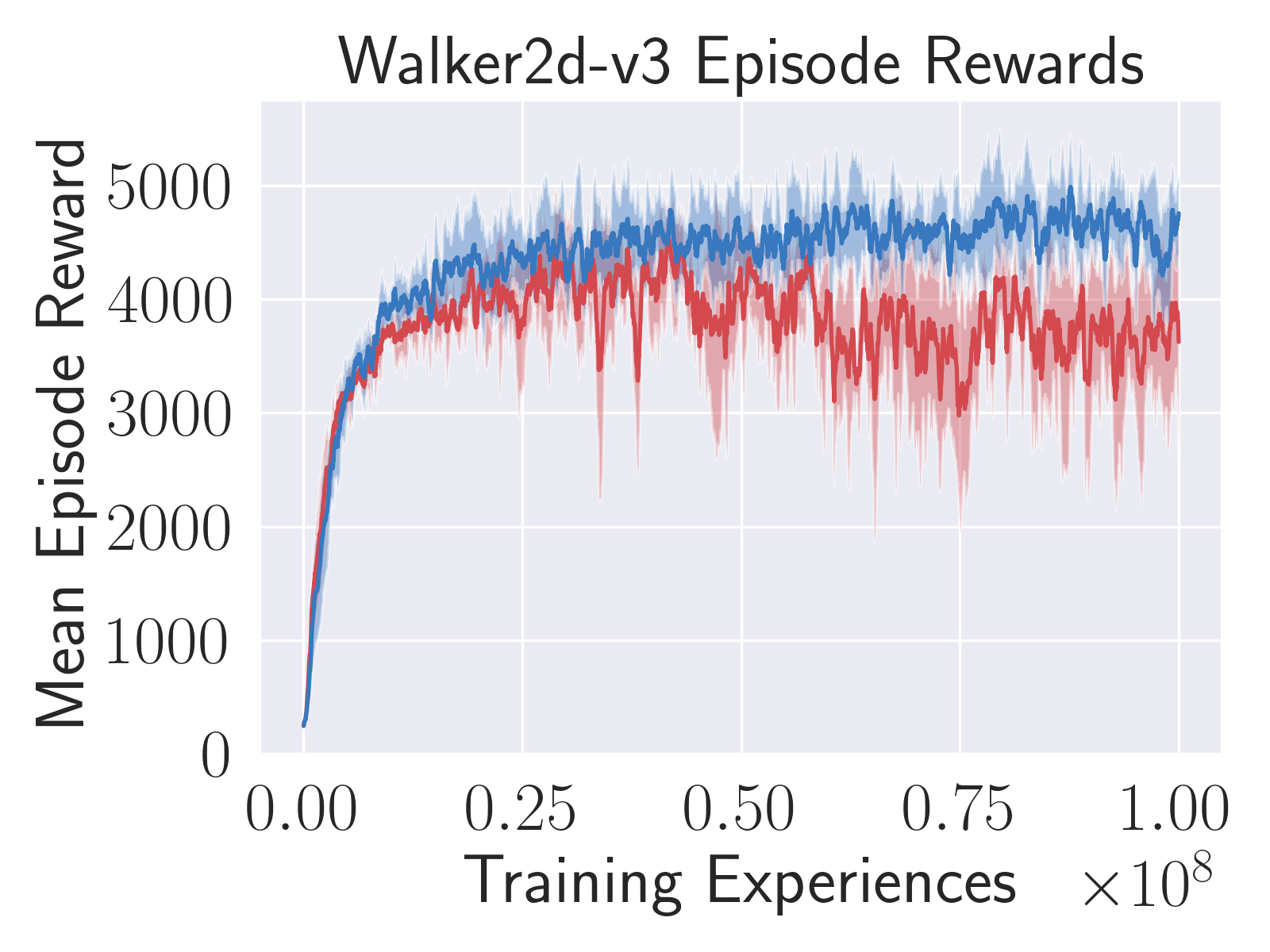}
\includegraphics[width=0.24\textwidth]{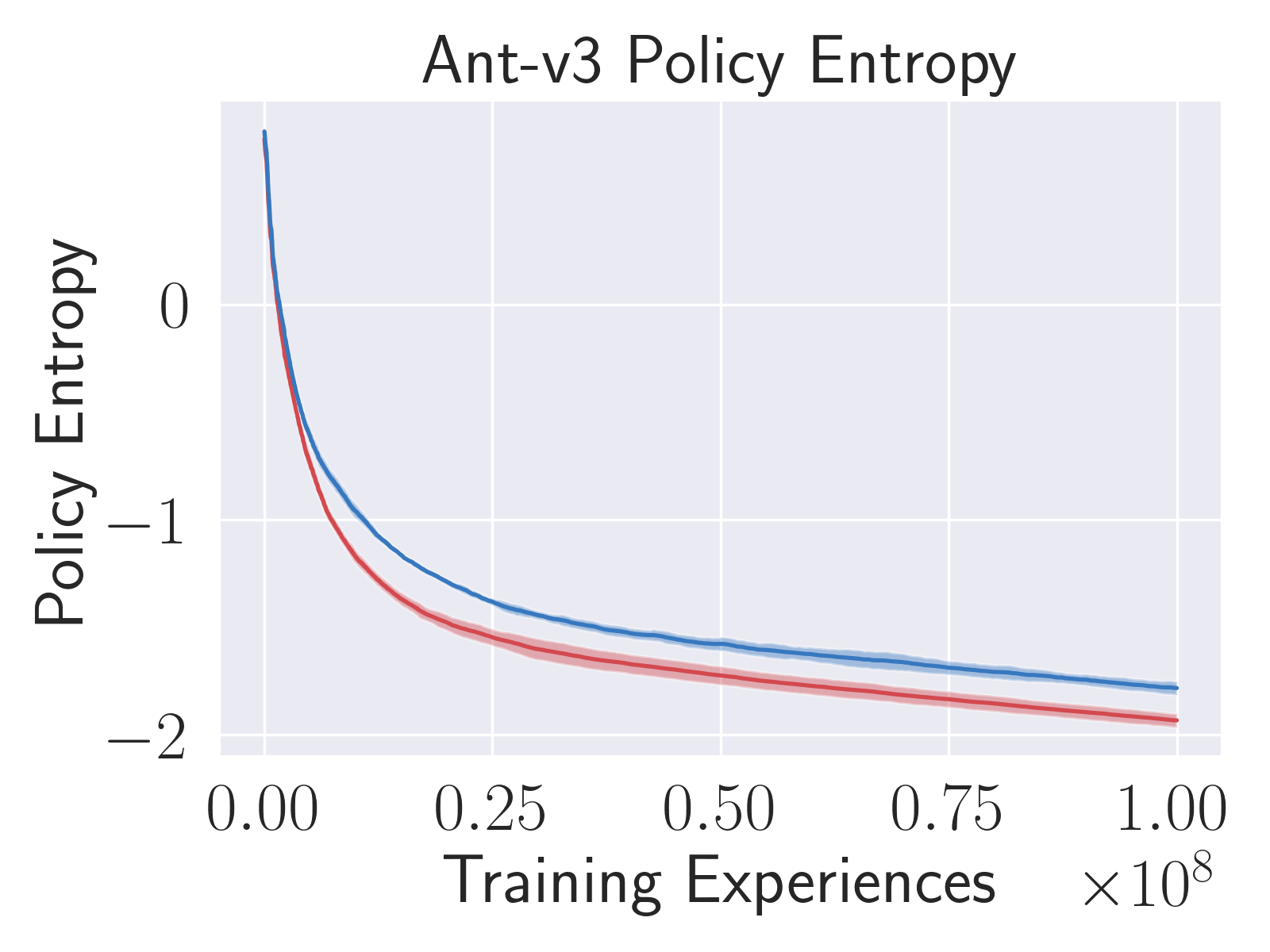}
\includegraphics[width=0.24\textwidth]{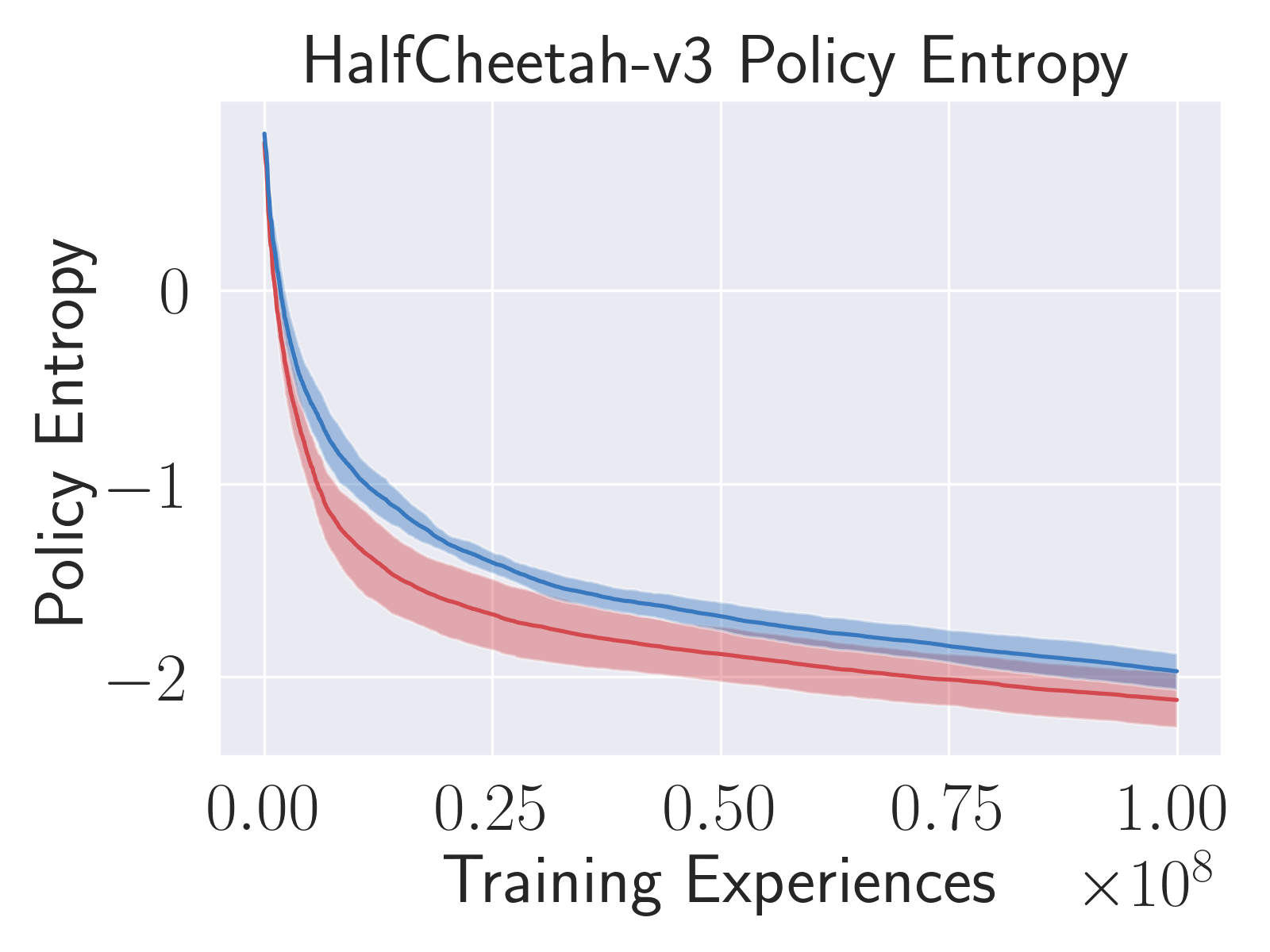}
\includegraphics[width=0.24\textwidth]{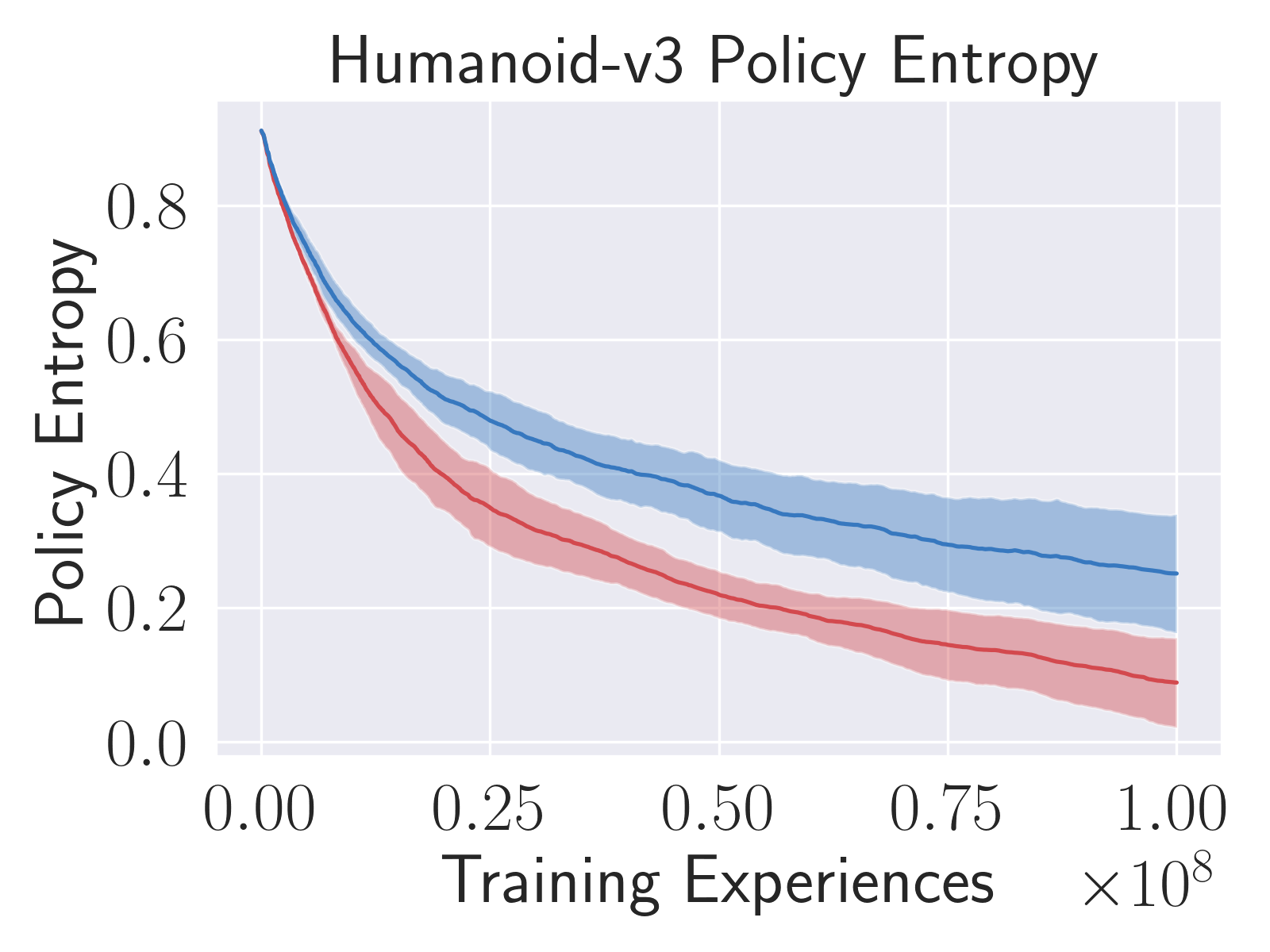}
\includegraphics[width=0.24\textwidth]{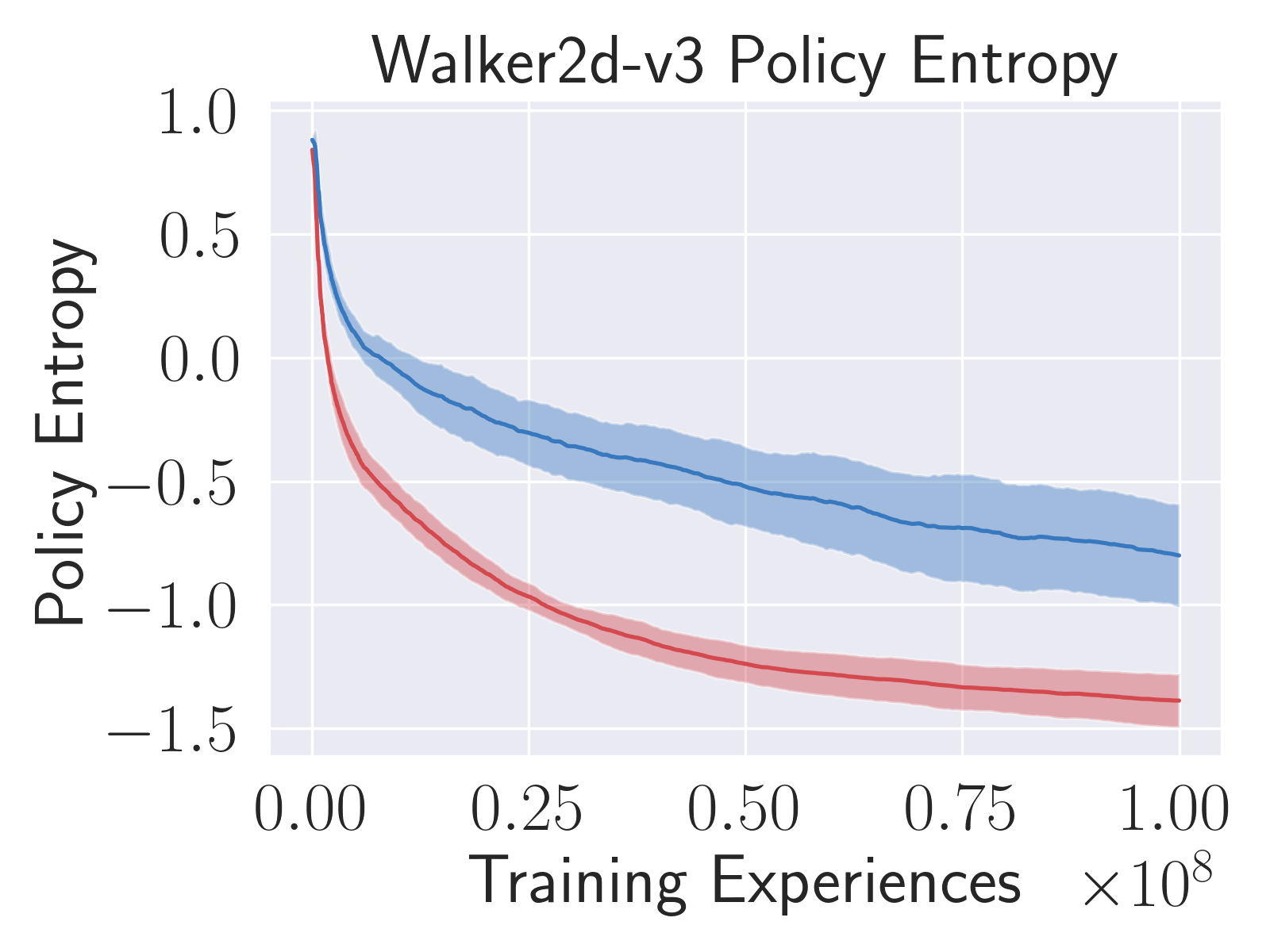}
\includegraphics[width=0.9\textwidth]{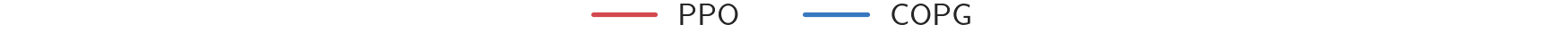}
\caption{Comparison of COPG (blue) and PPO (red) in MuJoCo environments.  Top row: COPG consistently produces higher final rewards and more stable training.  Bottom row:  COPG maintains a higher policy entropy throughout training.}\label{mujoco}
\end{figure}

We first pursued a direct comparison of the COPG objective with standard PPO in well-known MuJoCo environments, using the formulation, configuration, and hyperparameters of \cite{SpinningUp2018}.  The results are given in Figure \ref{mujoco}.  As expected, policy entropy is consistently higher throughout training with the clipped-objective policy gradient.  This is paired with improved learning from COPG, in terms of both reward levels and stability of the learning curve.  Notably, the learning of COPG does not require more samples than that of PPO, despite the relatively higher entropy its policies maintain.

\begin{figure}
\centering
\includegraphics[width=0.32\textwidth]{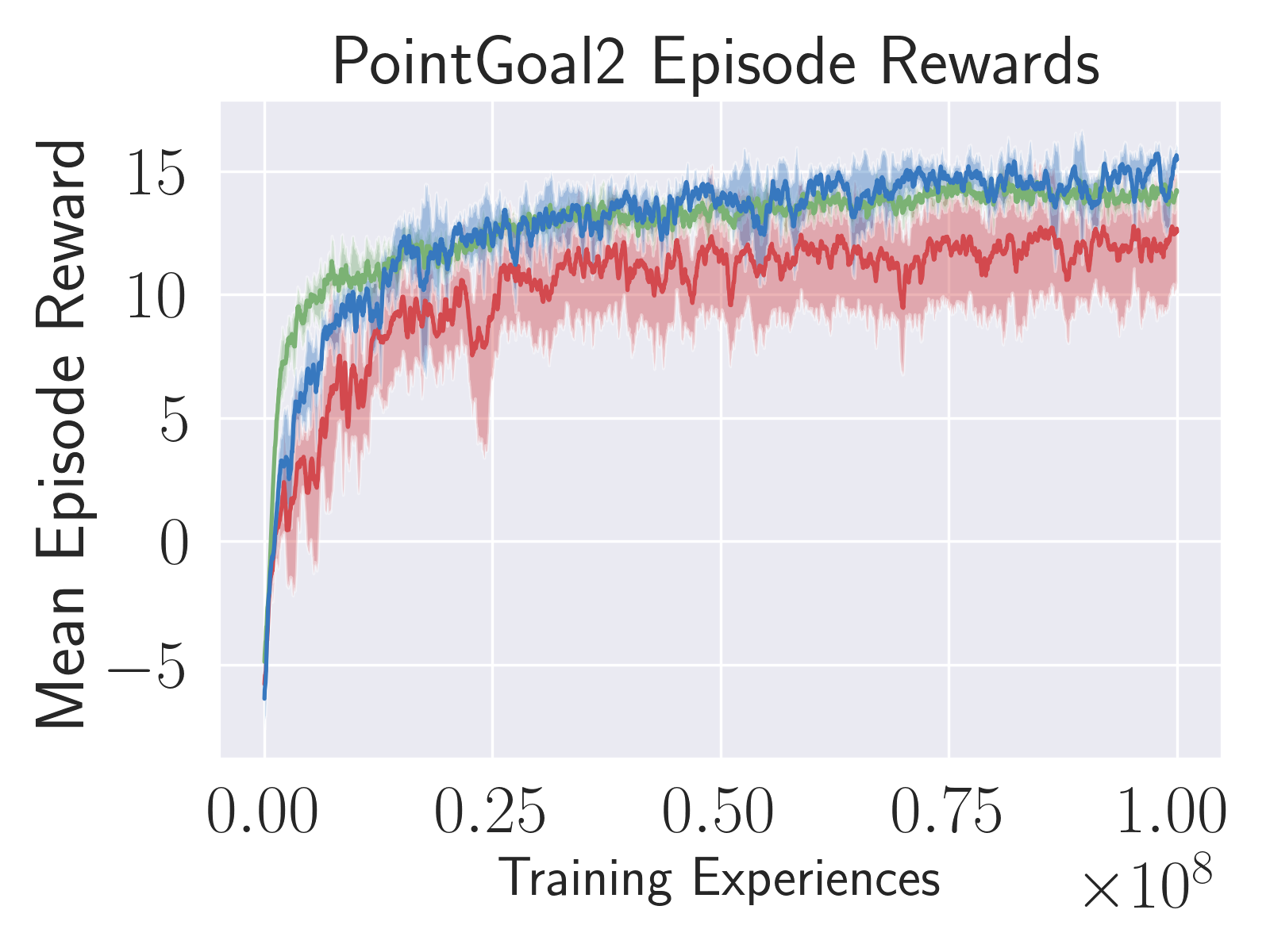}
\includegraphics[width=0.32\textwidth]{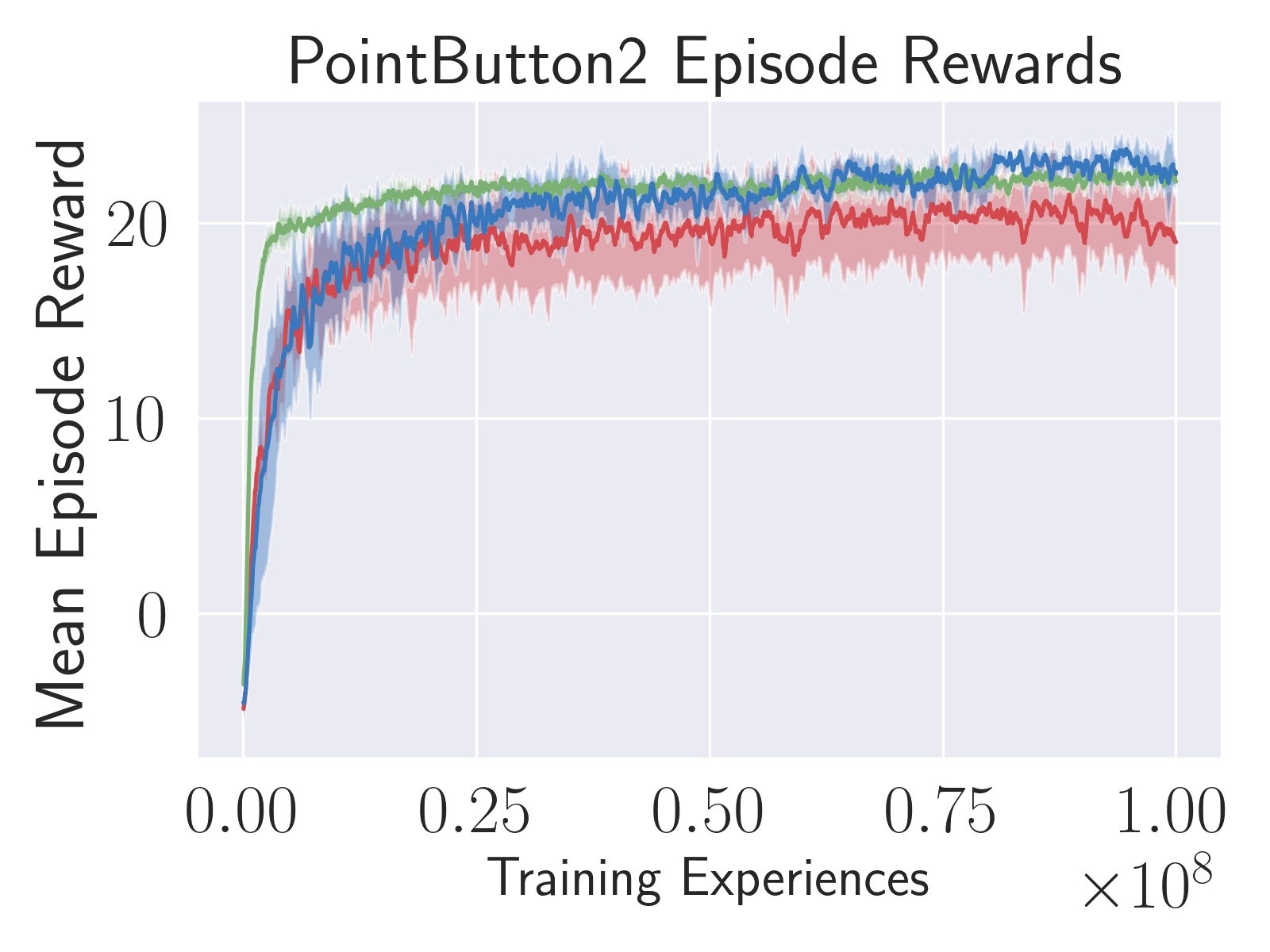}
\includegraphics[width=0.32\textwidth]{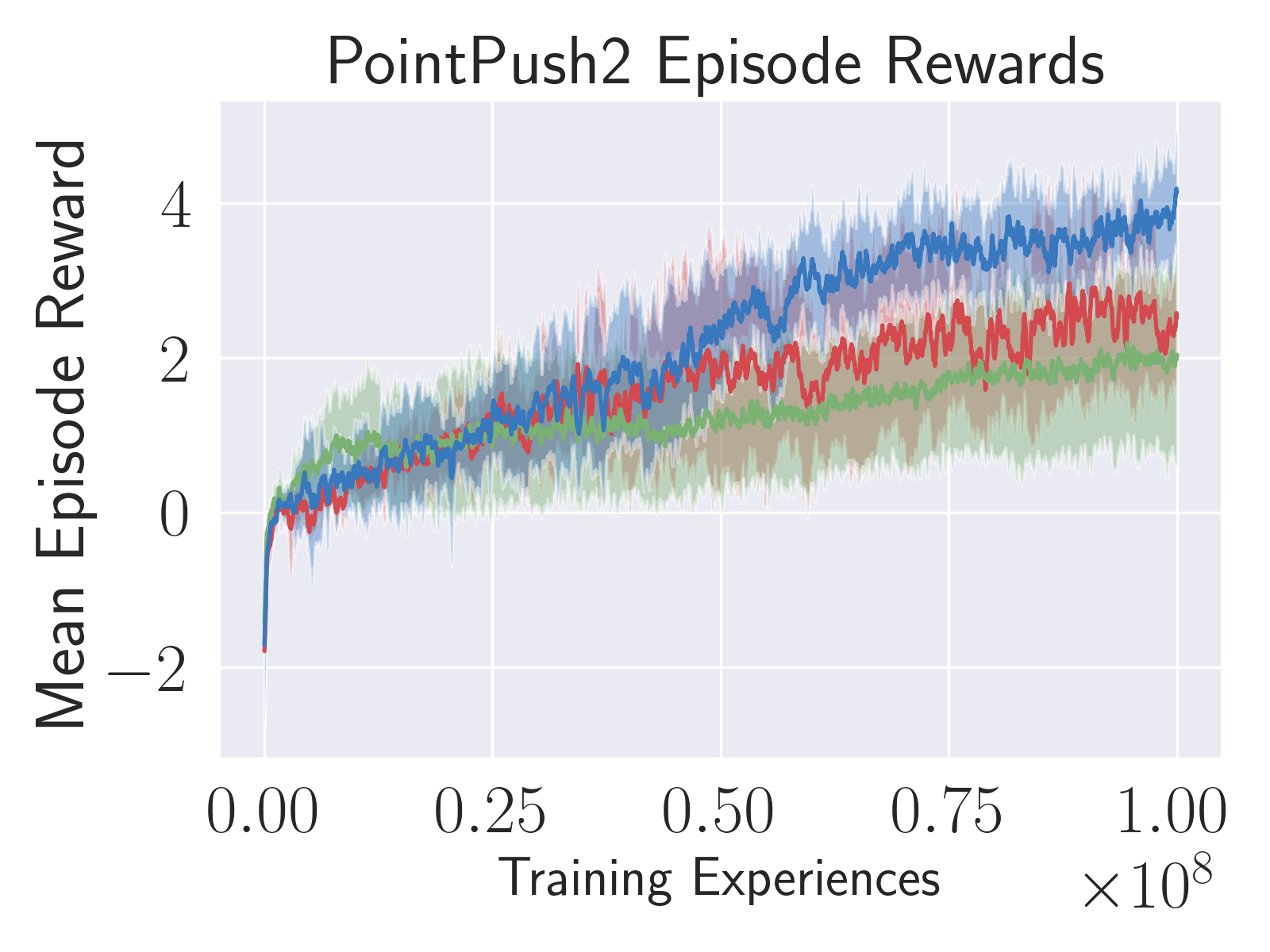}
\includegraphics[width=0.9\textwidth]{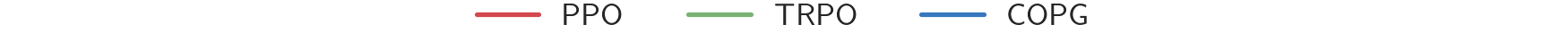}
\caption{COPG, PPO, and TRPO are compared without constraints on three Safety Gym environments with the ``Point'' robot.  COPG consistently outperforms PPO and is comparable or slightly superior to TRPO, while retaining the simplicity of a first-order method.}\label{safety_gym_1}
\end{figure}

To evaluate the impact of the clipped-objective policy gradient in more stochastic environments, we used the OpenAI Safety Gym \cite{RaAcAm19}. Safety Gym is a configurable suite of robotic navigation tasks that require continuous, multidimensional control.  It contains multiple types of robots that must navigate around obstacles with different dynamics in order to perform different tasks.  Accordingly, it contains both positive and negative reward terms and allows for evaluation of how agents handle risk and constraints.  Safety Gym is highly stochastic, as the locations of the obstacles and goals are randomized at the start of each episode.

Safety Gym tracks adverse events but does not include them in the reward function. In our next set of (unconstrained) experiments, we assigned each adverse event a fixed, negative reward.  In the constrained setting, which we discuss subsequently, the coefficient for cost events is learned.  To challenge our agents and highlight outcome variability, we focused on the most obstacle-rich (level 2) publicly available environments. In particular, we evaluated our methods on the ``Point'' and ``Car'' robots on each task (``Goal”, ``Button”, and ``Push”).  We used hyperparameters identical to the baseline experiments in \cite{RaAcAm19}, and included the clipped-action policy gradient of \cite{pmlr-v80-fujita18a} in all tested algorithms.  Further discussion of our Safety Gym setup, including penalty weights for the unconstrained experiments, is provided in Appendix \ref{safety_gym_app}.

The results of our unconstrained runs using the Point robot are shown in Figure \ref{safety_gym_1}, while the results for the Car robot are deferred to Appendix \ref{results_app}.  In all cases, the clipped-objective policy gradient outperforms PPO with its standard (importance sampling) objective.  COPG performance is comparable to or slightly superior to that of TRPO, while being simpler to implement.  We also observe that policy entropy is typically higher with COPG than with PPO, throughout training (Appendix \ref{results_app}).  This is true whether or not control bounds are taken into account in the entropy calculation.

\begin{figure}
\centering
\includegraphics[width=0.32\textwidth]{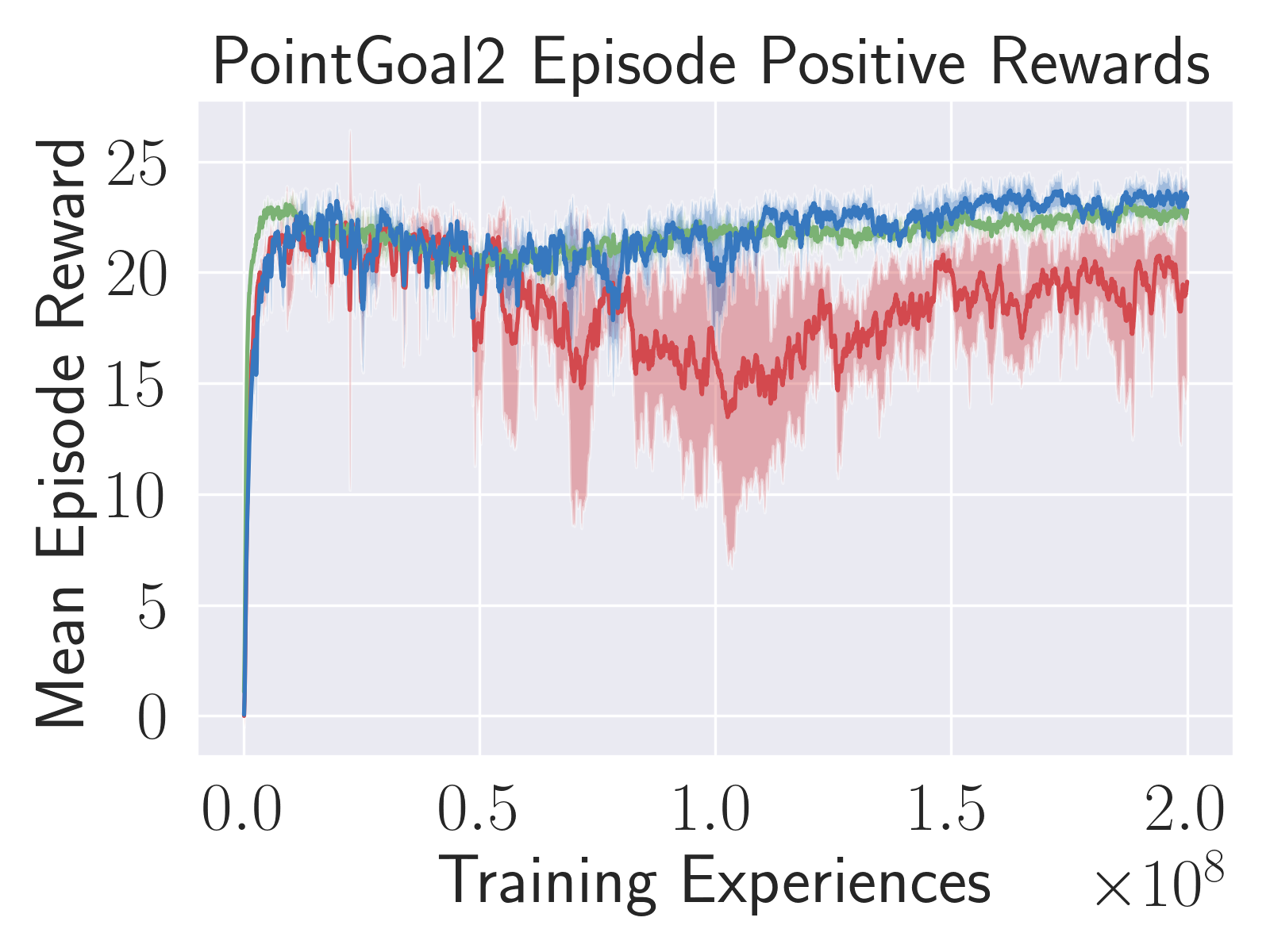}
\includegraphics[width=0.32\textwidth]{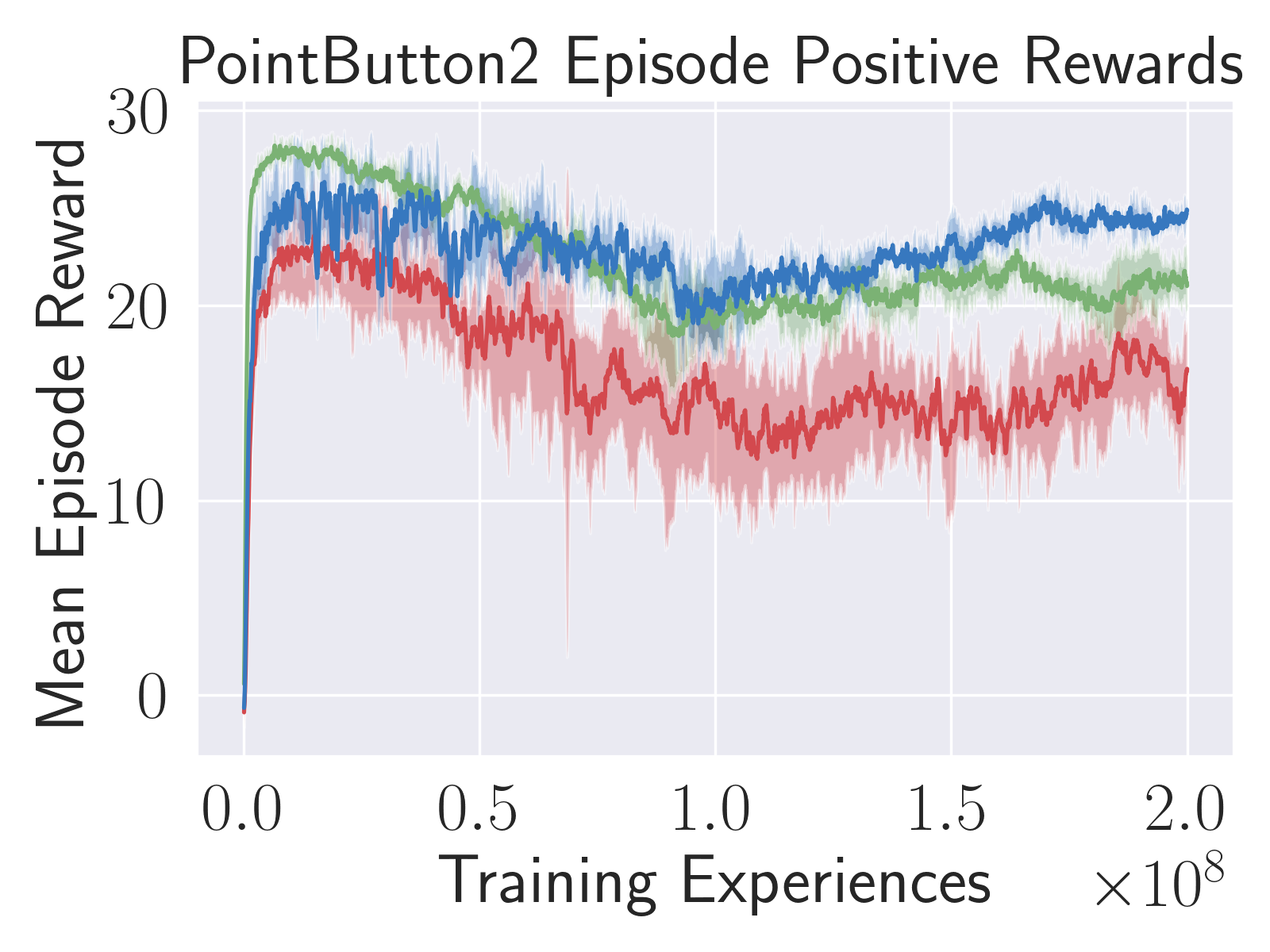}
\includegraphics[width=0.32\textwidth]{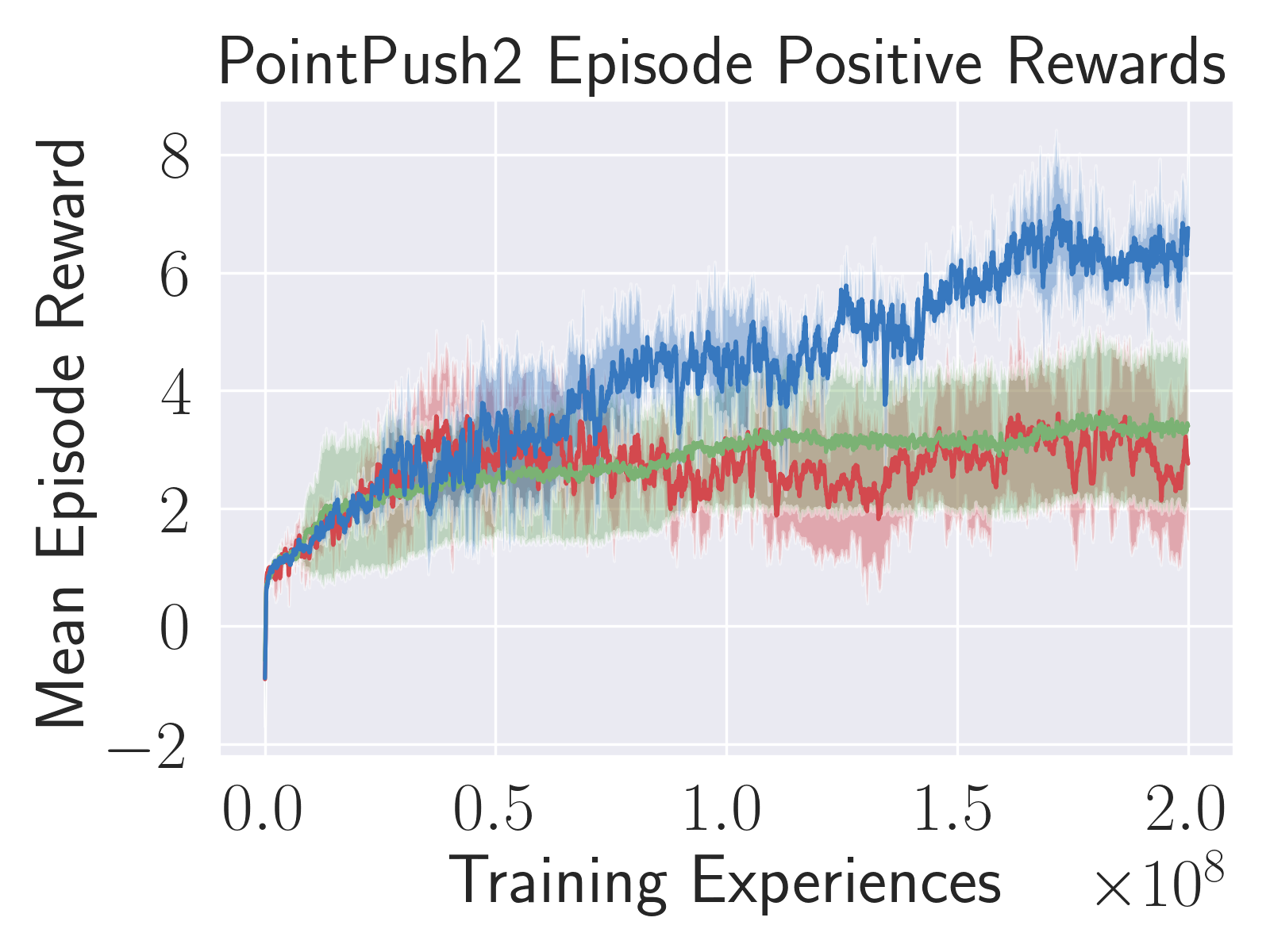}
\includegraphics[width=0.32\textwidth]{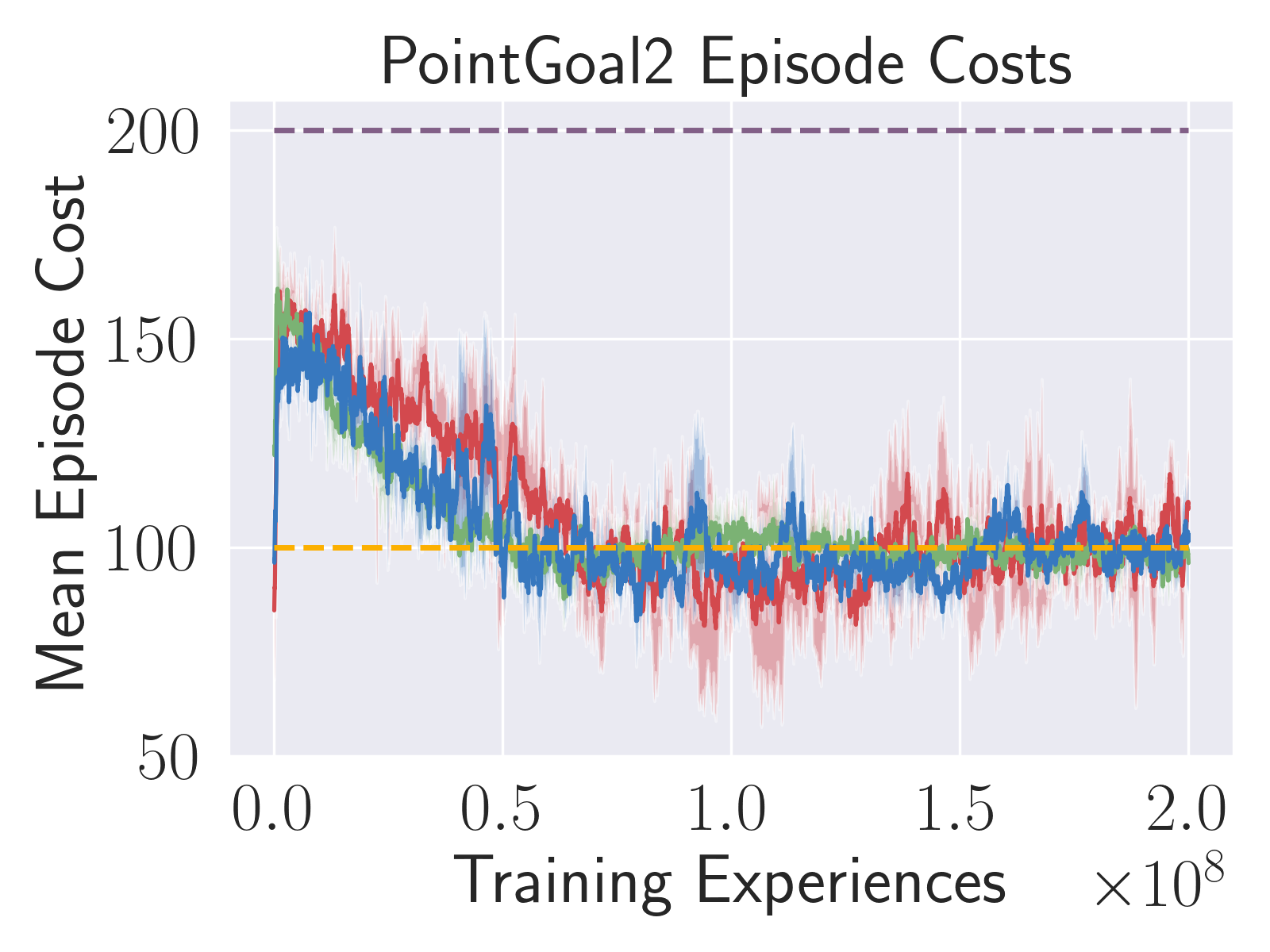}
\includegraphics[width=0.32\textwidth]{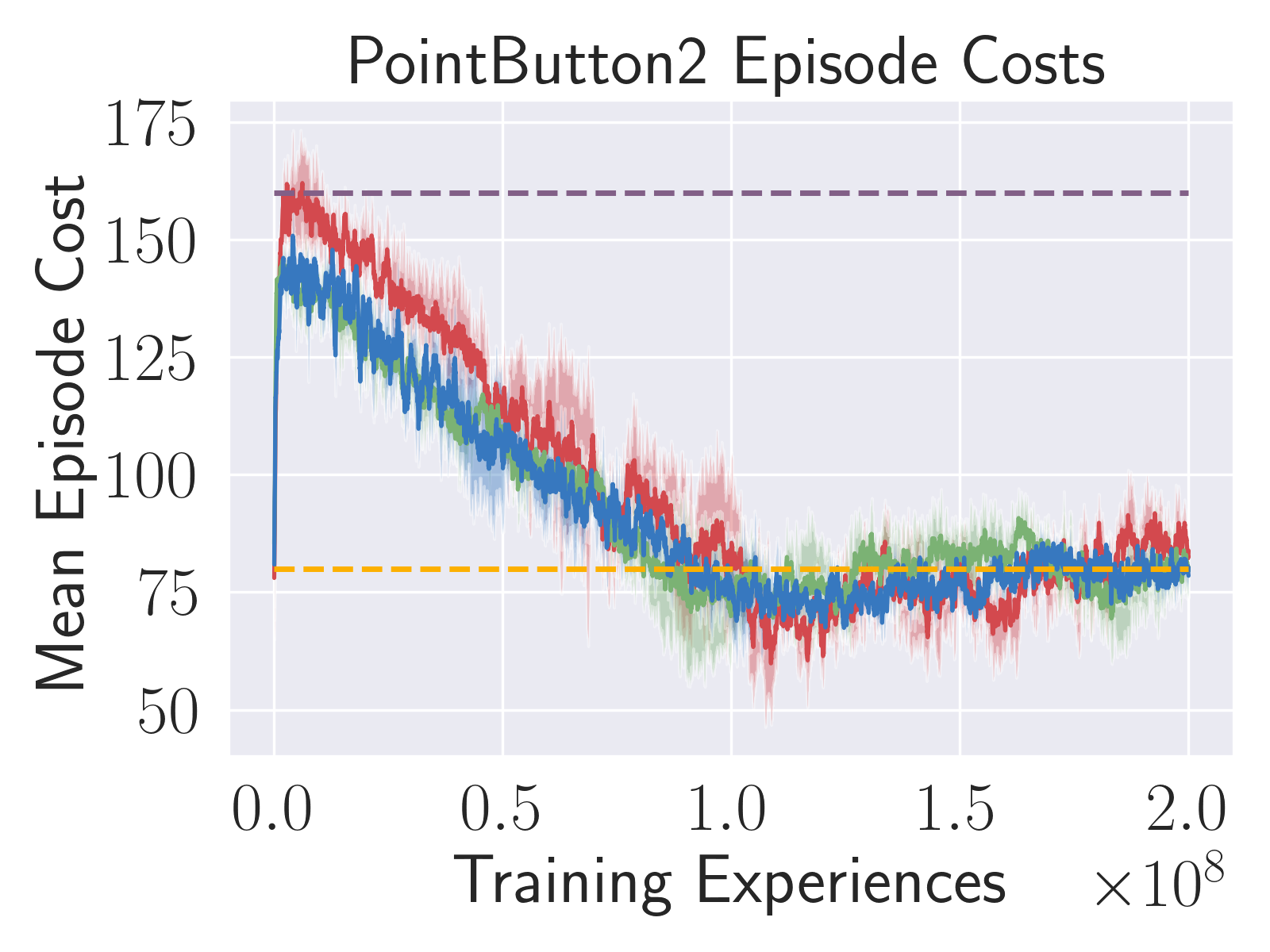}
\includegraphics[width=0.32\textwidth]{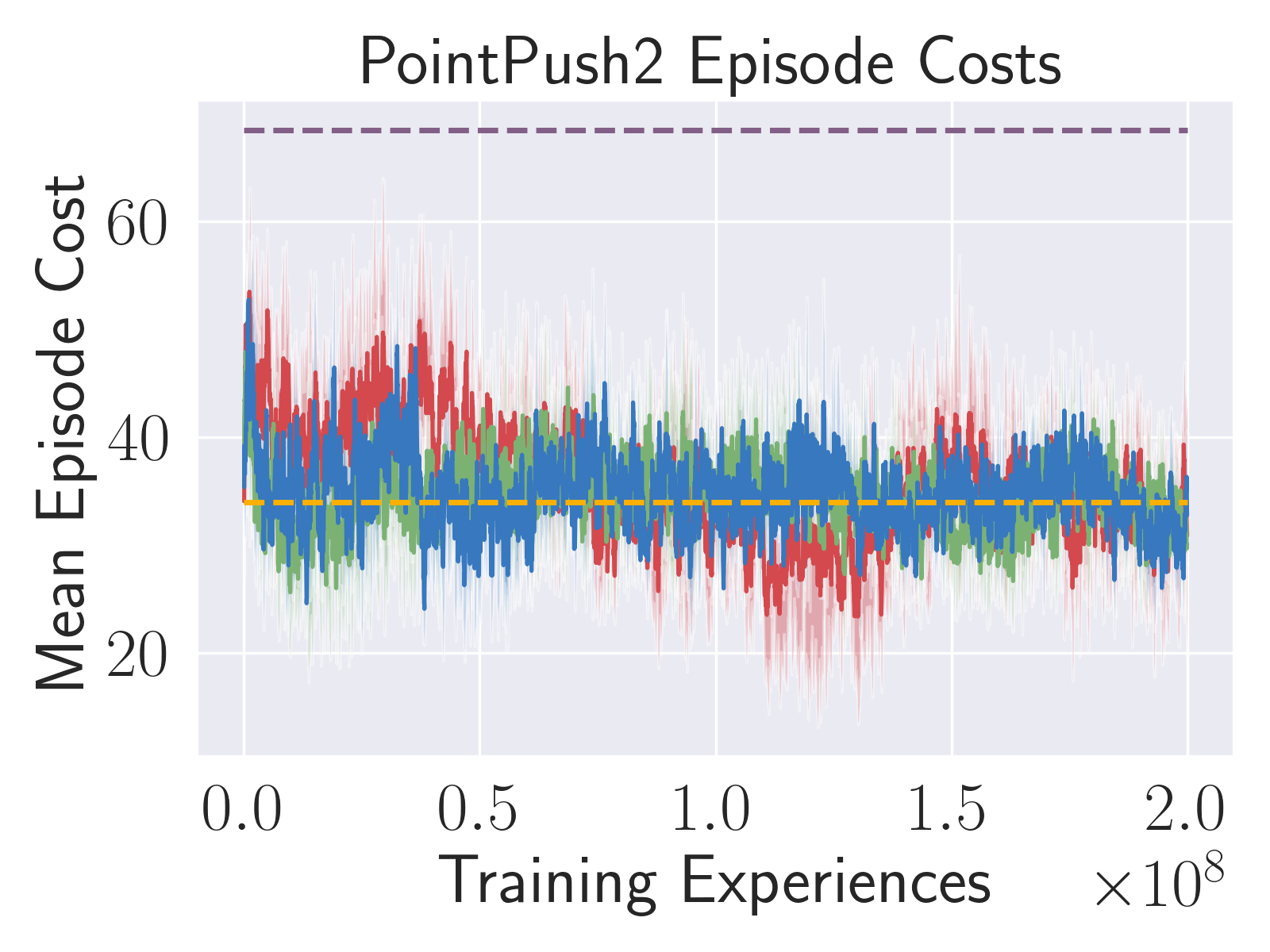}
\includegraphics[width=0.9\textwidth]{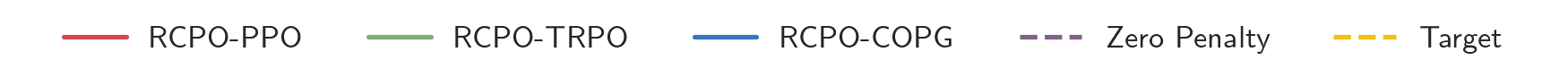}
\caption{Comparison of RCPO constrained policy optimization driven by PPO, TRPO, and COPG for Point robot environments. RCPO COPG obtains higher positive rewards while converging to the target cost level as quickly as the others.} \label{safety_gym_2}
\end{figure}

\begin{figure}[H]
    \centering
    \includegraphics[width=0.32\textwidth]{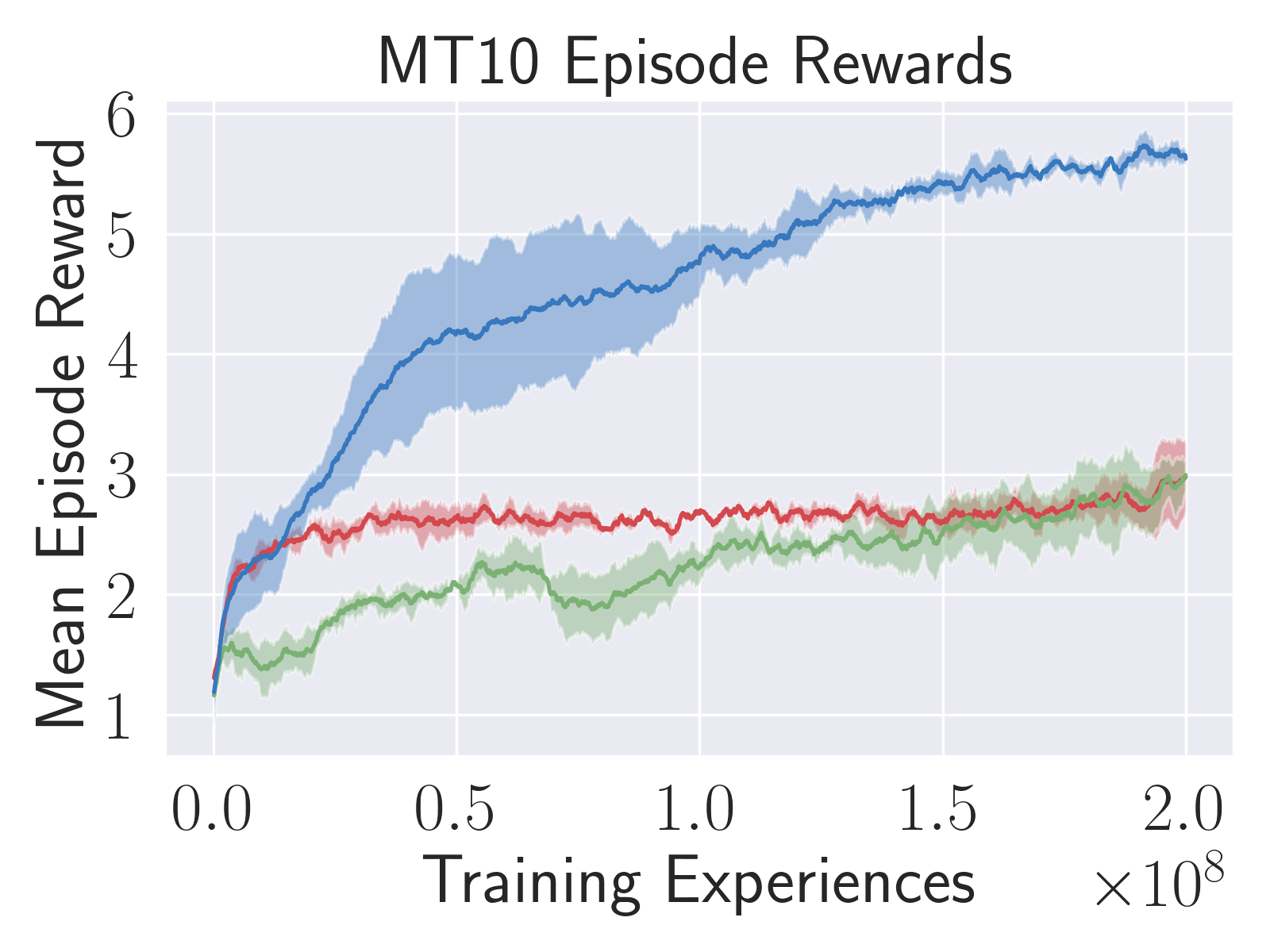}
    \includegraphics[width=0.32\textwidth]{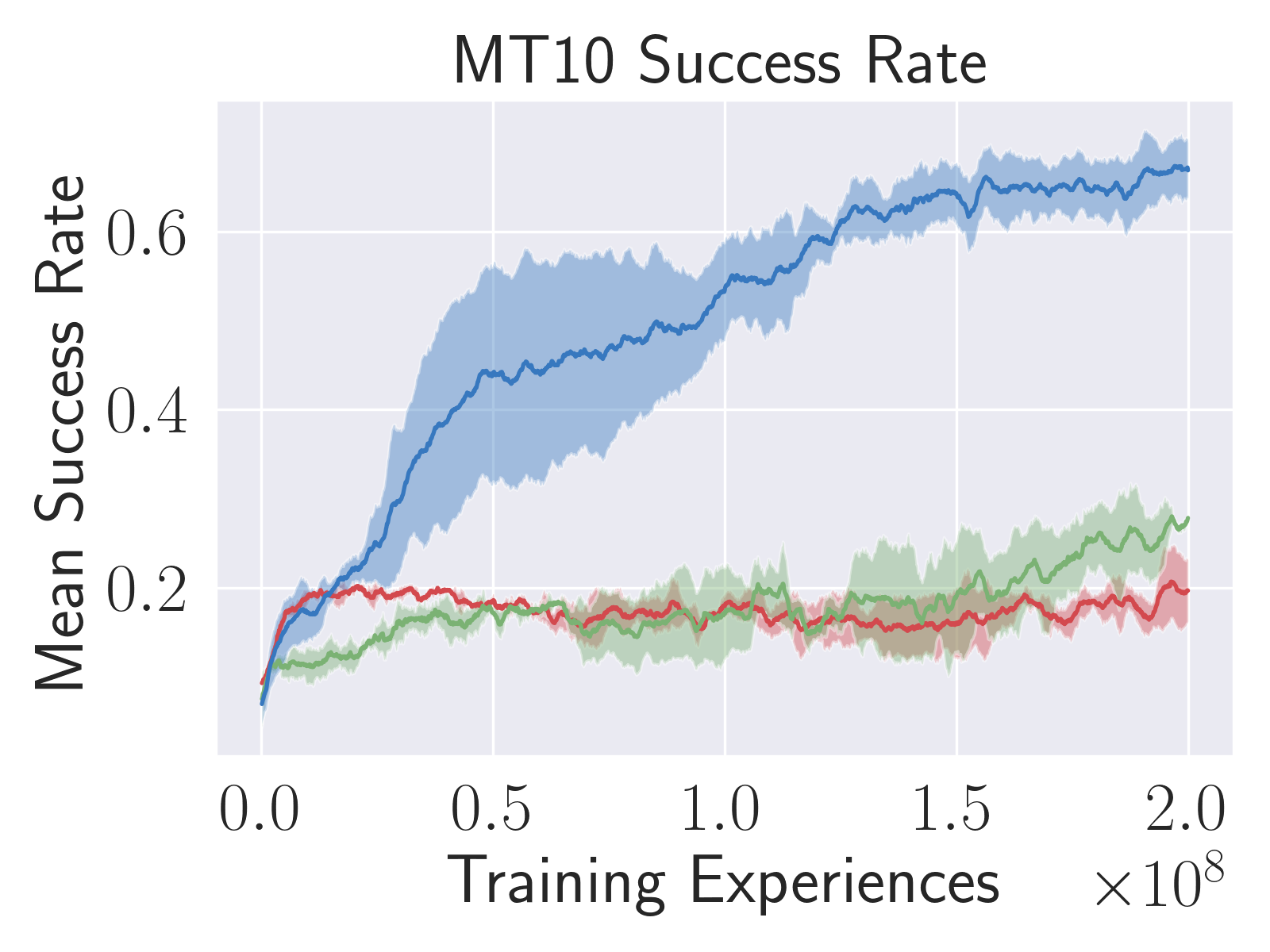}\\
    \includegraphics[width=0.32\textwidth]{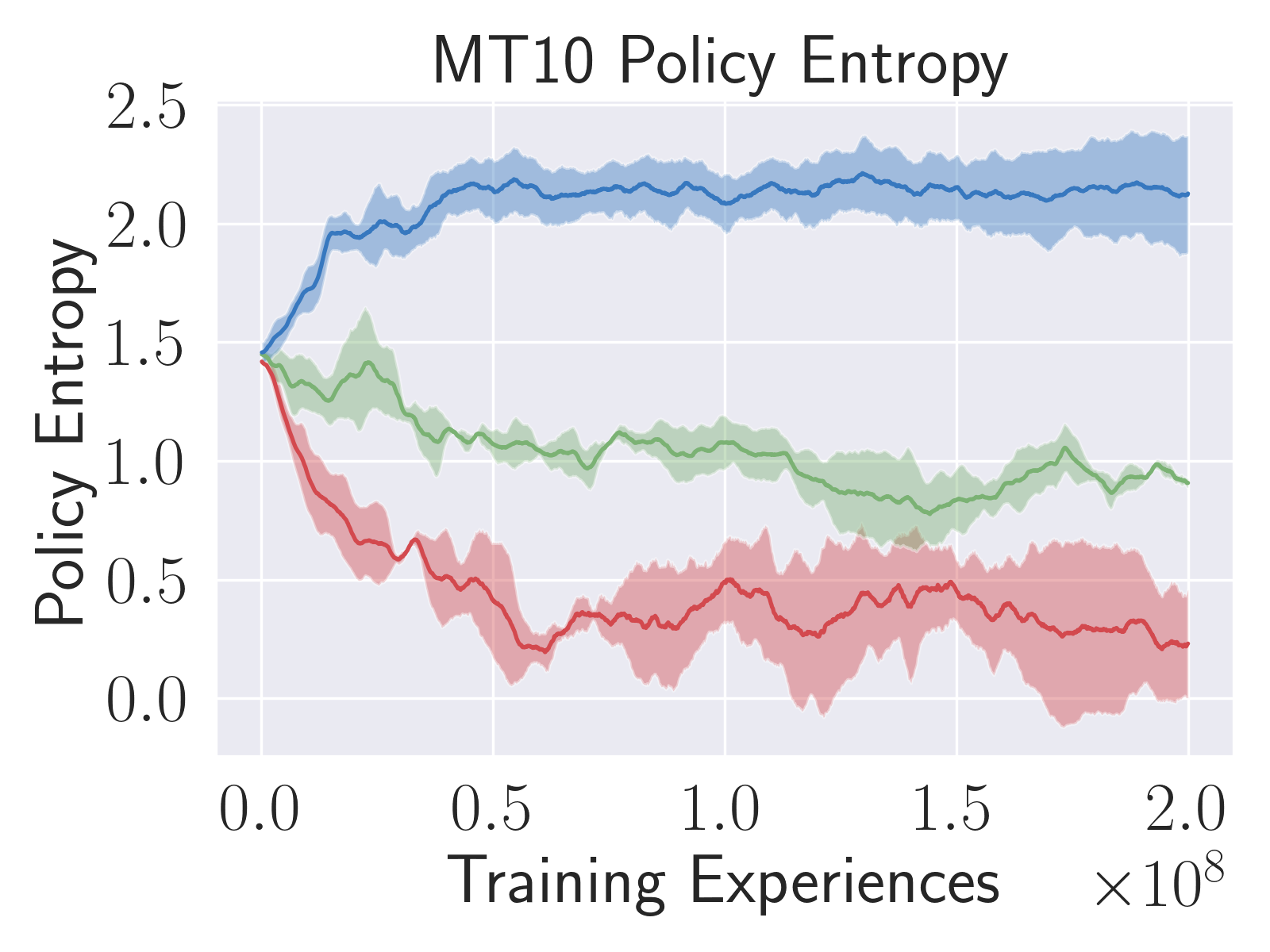}
    \includegraphics[width=0.32\textwidth]{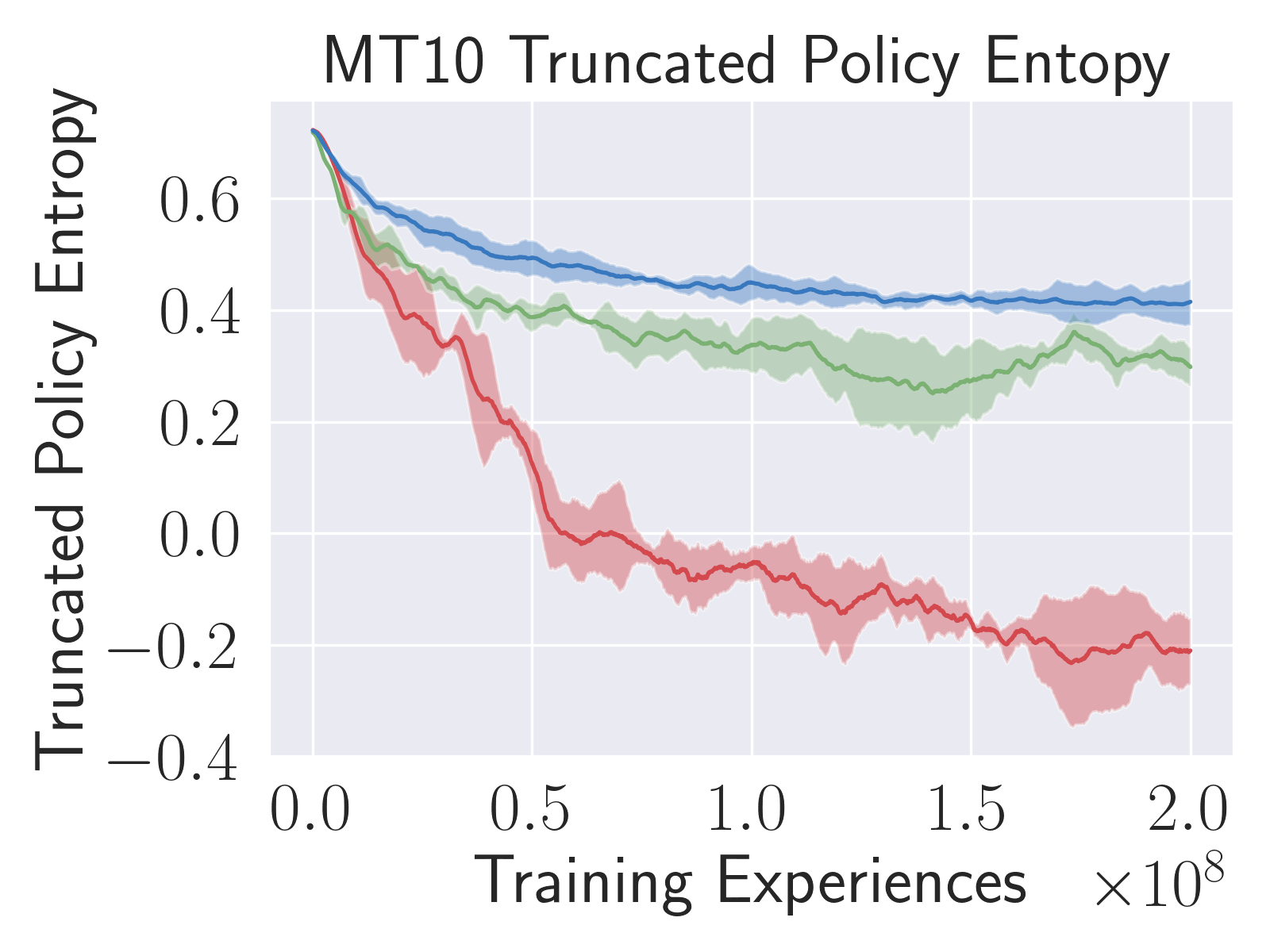}\\
    \includegraphics[width=0.9\textwidth]{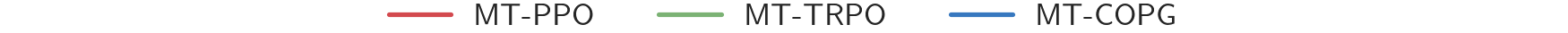}
    \caption{Comparison of PPO, TRPO, and the COPG on the Meta-World MT10 benchmark.  Top row: the COPG significantly outperforms others, both in terms of reward accumulation and success rate.  Bottom row: the COPG maintains higher policy entropy than the other methods, both considering control bounds (right) and not (left).}
    \label{mt10}.
\end{figure}

\subsection{Constrained, Single-Task Learning}
The clipped-objective policy gradient can also be applied to the constrained setting, in the same way that PPO is.  In Constrained Markov Decision Processes, positive rewards $r(\mathbf{s}_t, \mathbf{a}_t)$ and costs $c(\mathbf{s}_t, \mathbf{a}_t)$ are defined for each time step, and a constraint $C(\tau) = F(c(\mathbf{s}_1, \mathbf{a}_1), \ldots, c(\mathbf{s}_T, \mathbf{a}_T))$ is defined over each trajectory.  The associated learning problem is then to find
\begin{equation}
    \max_\theta J_R(\theta) \text{ s.t. } J_C(\theta) \le d,
\end{equation}
where $J_R(\theta)$ is the objective based on positive reward, $J_C(\theta) = E_{\tau \sim p_\theta(\tau)}C(\tau)$, and $d$ is a fixed threshold.

The most common approach to finding an optimal policy with such a constraint is to concurrently learn a scaling factor for cost relative to positive reward, via dual optimization. As in \cite{arxiv_version}, we integrate the PPO, TRPO, and COPG objectives into Reward Constrained Policy Optimization (RCPO; \cite{TeMaMa19}), a recent algorithm for learning constrained policies.  

We evaluate RCPO driven by these algorithms on the same set of Safety Gym environments, in each case choosing the cost target to be half of the cost accumulated by an unconstrained agent trained without knowledge of the cost terms.  The results for the Point robots are shown in Figure \ref{safety_gym_2}, while the results for the Car robots are again deferred to Appendix \ref{results_app}.  The use of the COPG in RCPO compares favorably with the use of both PPO and TRPO in terms of the accumulation of positive reward, while reaching the target cost level at approximately the same rate.

\subsection{Multi-Task Learning}
Finally, we applied the clipped objective policy gradient to a challenging multi-task learning problem; the MT10 benchmark of the Meta-World suite \cite{metaworld}. Meta-World is comprised of a set of manipulation tasks to be performed by a Sawyer robotic arm.  Each task has 50 variations (for instance in the placement of a goal object); this means that the MT10 benchmark has 10 broad tasks with 50 variations each.

We compared the performance of the clipped-objective policy gradient on MT10 with that of TRPO and PPO, using the same hyperparameters as given in \cite{metaworld}.  The lone difference in the configuration compared to \cite{metaworld} was that the bound correction of \cite{pmlr-v80-fujita18a} was applied in each method.  We observed that the COPG approach substantially outperformed the others, both in terms of reward and success rate (Figure \ref{mt10}).  Its final performance, while not achieved in as few samples, is on par with or slightly exceeds that of Soft Actor-Critic \cite{sac} on the benchmark\cite{metaworld}.  The strong performance of the COPG is likely derived from its ability to improve while exploring more thoroughly than PPO or TRPO, as evidenced by the observed differences in policy entropy (Figure \ref{mt10}, second row).

\section{Discussion}
One strategy for improving the sample efficiency of policy optimization is to take multiple gradient steps per batch of collected data.  Proximal Policy Optimization is the most popular algorithm that follows this approach.  We propose an alternative that maintains everything about PPO (configuration, hyperparameters, etc.), with the exception of its objective.  Policy gradient estimates from both PPO and COPG are biased; however, they have much lower variance than those of the unbiased off-policy policy gradient.  In particular, the COPG approach essentially ignores the difference between the policy under which a batch was collected and the policy that is currently being updated, provided their outputs are sufficiently close.

As shown through our results, the tendency of COPG to provide smaller updates toward regions of positive advantage and larger updates away from regions of negative advantage leads to learning that is consistently superior to that of PPO for continuous action spaces.  While these gains differ in structure from those that account for the ``primacy bias'' in off-policy learning \cite{primacy}, a similar intuition is at play: we seek to prevent the agent from prematurely converging to a suboptimal maximum.  This motivation is similarly present in max-entropy RL strategies (e.g., \cite{sac}).  

The ability of the COPG to maintain high policy entropy throughout training suggests that it may be particularly impactful in environments where exploration is key.  This intuition is supported by its relatively strong performance in multi-task learning, though more investigation is required.  Also interesting would be to evaluate the performance of the method on discrete action spaces, particularly those with many possible actions and that require significant exploration.  In particular, the method may compare favorably to PPO when applied to large language models (LLMs) via reinforcement learning from human feedback (RLHF; e.g., \cite{rlhf}). 

Given the observed performance of the COPG, PPO, and TRPO, as well as the simplicity and compute requirements of the COPG (which essentially match those of PPO), we believe the COPG can and should be frequently applied to learning problems with continuous action spaces.  While it may not be as sample efficient as off-policy methods (notably soft actor-critic \cite{sac}), its performance, stability, simplicity, and ability to be naturally parallelized make it a viable alternative.  We also note that the COPG pairs naturally with risk-sensitive approaches, as shown in \cite{arxiv_version}.

\section{Conclusions}
We propose and evaluate the clipped-objective policy gradient as an alternative to the importance sampling objective used in PPO.  We find that the relative ``pessimism'' of the clipped-objective policy gradient leads to enhanced exploration and learning that is consistently superior to that of PPO, for continuous action spaces.  The COPG was also found to provide performance that compares favorably with TRPO, while being simpler to implement.

\section{Acknowledgements}
The authors thank Cash Costello and I-Jeng Wang for insightful discussions of related topics.

\bibliographystyle{neurips}
\bibliography{main}

\section{Appendix} 

\subsection{Safety Gym Configuration} \label{safety_gym_app}
The choice to use the OpenAI Safety Gym \cite{RaAcAm19} to evaluate our approach stemmed from a desire to test in conditions with clear cost-benefit trade-offs, significant stochasticity, and adequate complexity.

The six environments chosen were the most obstacle-rich of the publicly-available environments that used either the ``Point'' or ``Car'' robot.  The Point robot lives in the 2D plane and has two control dimensions: one for moving forward/backward and one for turning. The Car robot also has two control dimensions, corresponding to two independently-actuated parallel wheels. It has a freely rotating wheel and typically remains in the 2D plaine (though it is not constrained to do so).  We expect our results to extend to the remaining default robot, ``Doggo,'' but did not attempt it because of the much longer training times it exhibited in \cite{RaAcAm19}.

We evaluated environments with several types of obstacles and tasks.  In all cases, the robot is given a fixed amount of time (1000 steps) to complete the given task as many times as possible. It is motivated by both sparse and dense reward terms.  In the ``Goal'' environments, the robot must navigate around obstacles to a series of randomly-assigned goal positions; as soon as a goal is reached, the robot is presented with a new goal.  In the ``Button'' environments, the robot must navigate to and press a sequence of goal buttons while avoiding obstacles and other buttons.  In the ``Push'' task, the robot must push a box around obstacles to a series of goal positions, again being presented with a new goal as soon as one is reached.  The set of obstacles are different for each task; among the three environments there are a total of five different constraint elements, each with different dynamics.  See \cite{RaAcAm19} for further details.

All of our experiments used the OpenAI default of a single indicator for overall cost at each time step.  In the unconstrained experiments, each cost event was incorporated into the reward function with a fixed (negative) weight.  This weight was 0.025, 0.05, and 0.075 in the Push, Button, and Goal environments, respectively.  These numbers were chosen to enable reasonable learning for the three different task types.

\subsection{Further Results}\label{results_app}
In addition to the experiments conducted with the Point robot that are shown in the main text, we performed both unconstrained and constrained trials with the Car robot.  Results are similar to those with Point and are provided here.  Plots of policy entropy throughout training are also provided for unconstrained learning in each environment, both with and without consideration of the control bounds. 

\begin{figure}[H]
\centering
\includegraphics[width=0.32\textwidth]{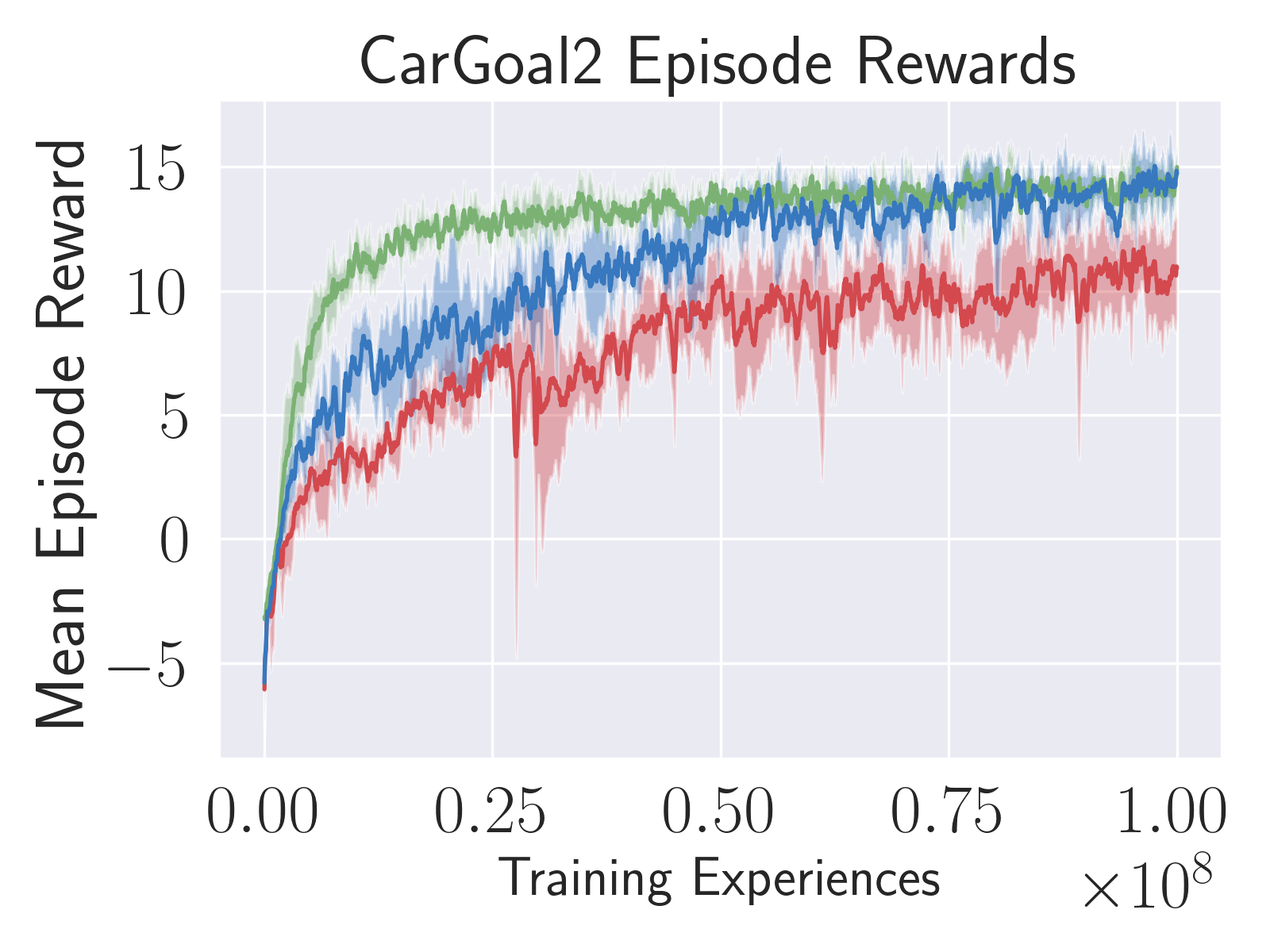}
\includegraphics[width=0.32\textwidth]{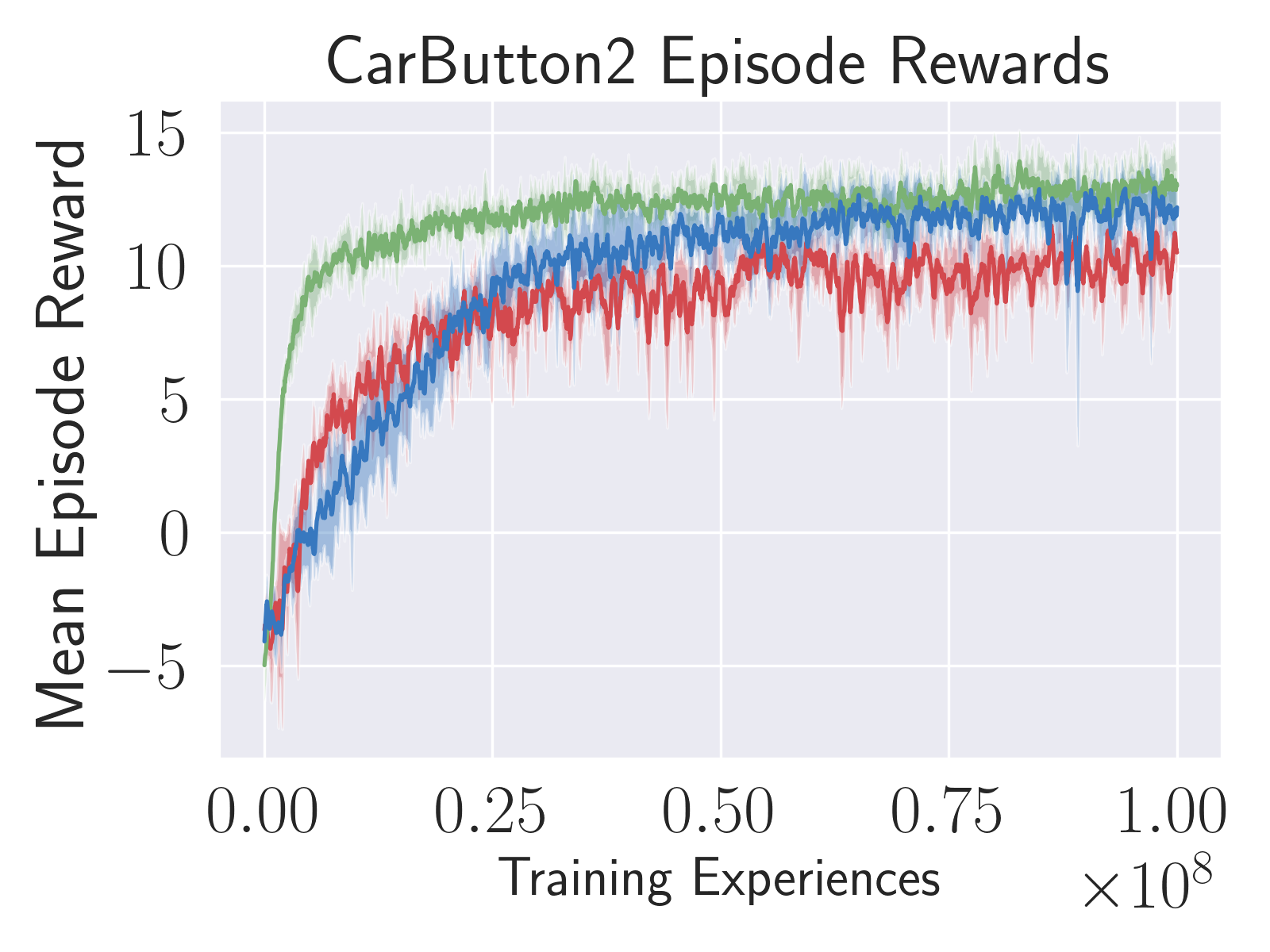}
\includegraphics[width=0.32\textwidth]{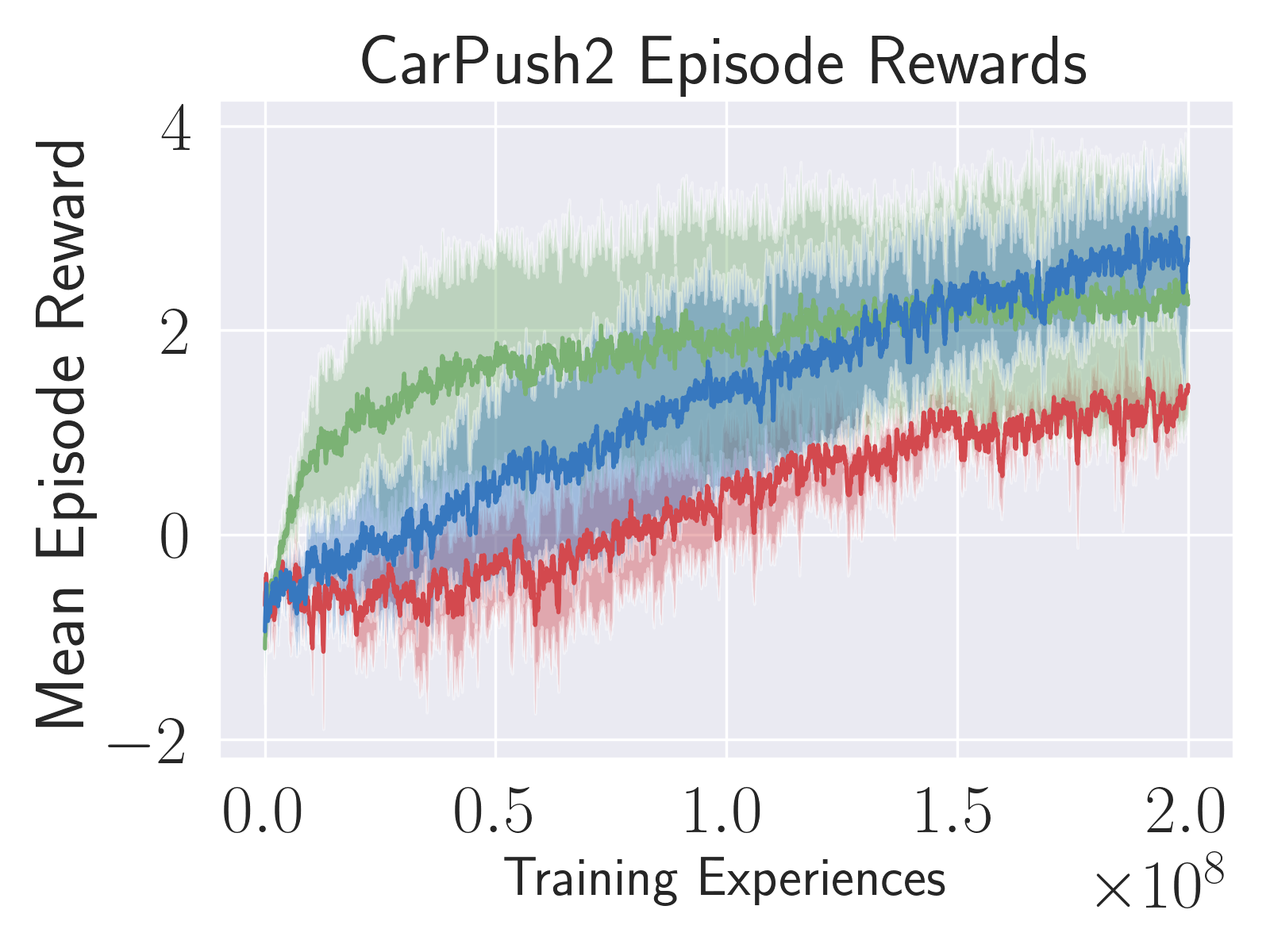}
\includegraphics[width=0.9\linewidth]{figures/legend_5.png}
\caption{COPG, PPO, and TRPO are compared on three Safety Gym environments without constraints and with the ``Car'' robot.  COPG consistently outperforms PPO and is comparable to TRPO.}
\end{figure}

\begin{figure}[H]
\centering
\includegraphics[width=0.32\textwidth]{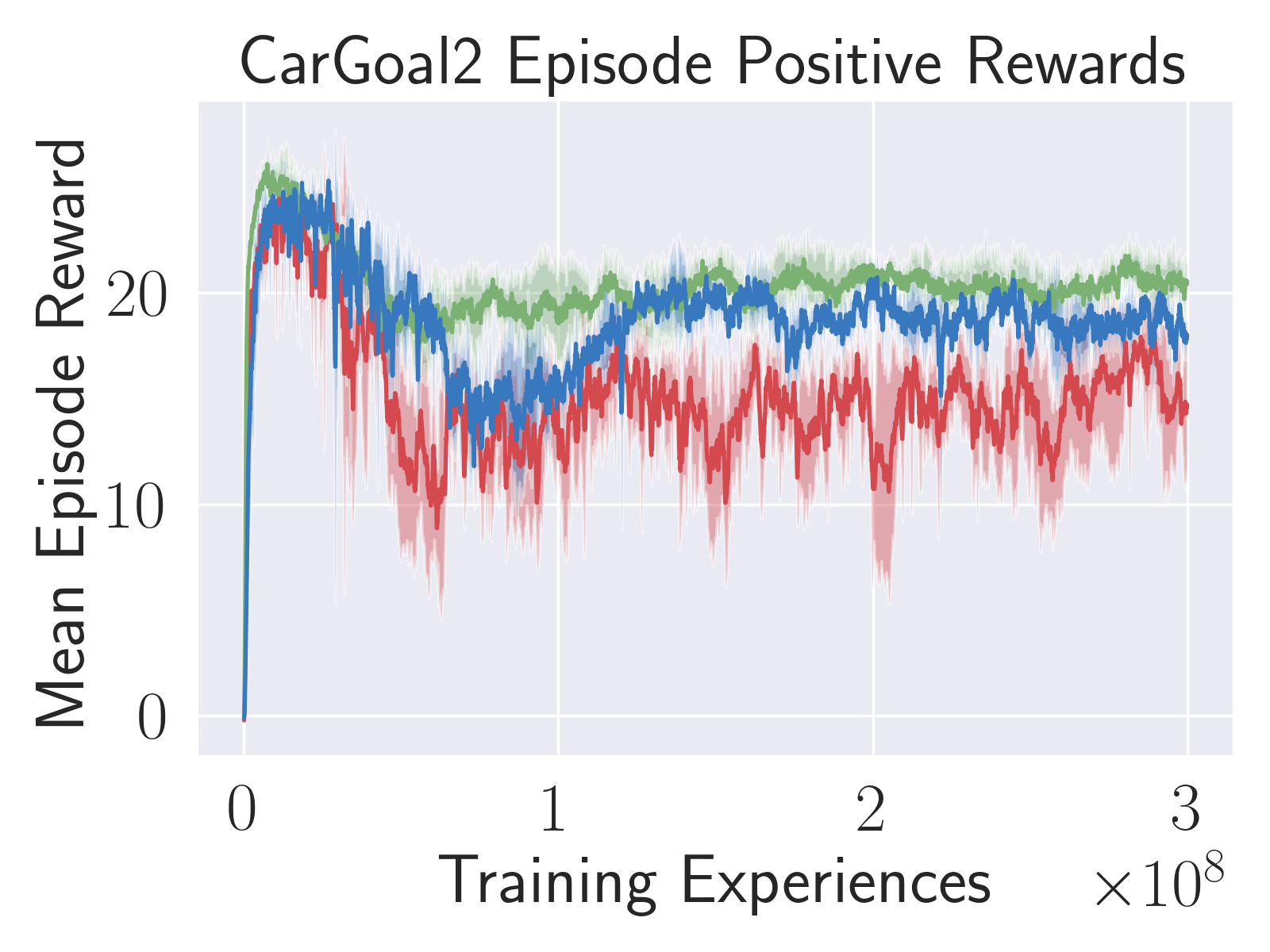}
\includegraphics[width=0.32\textwidth]{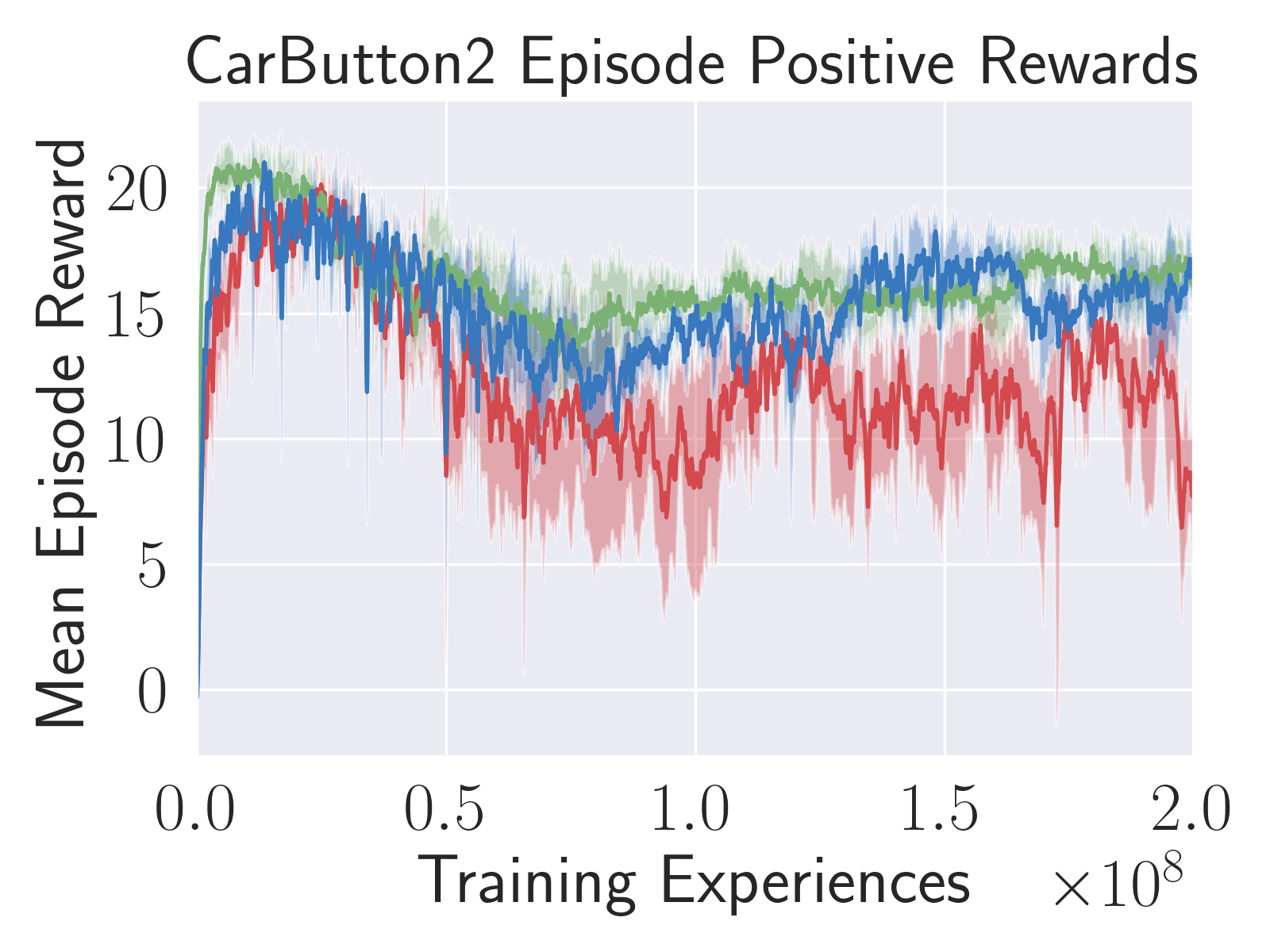}
\includegraphics[width=0.32\textwidth]{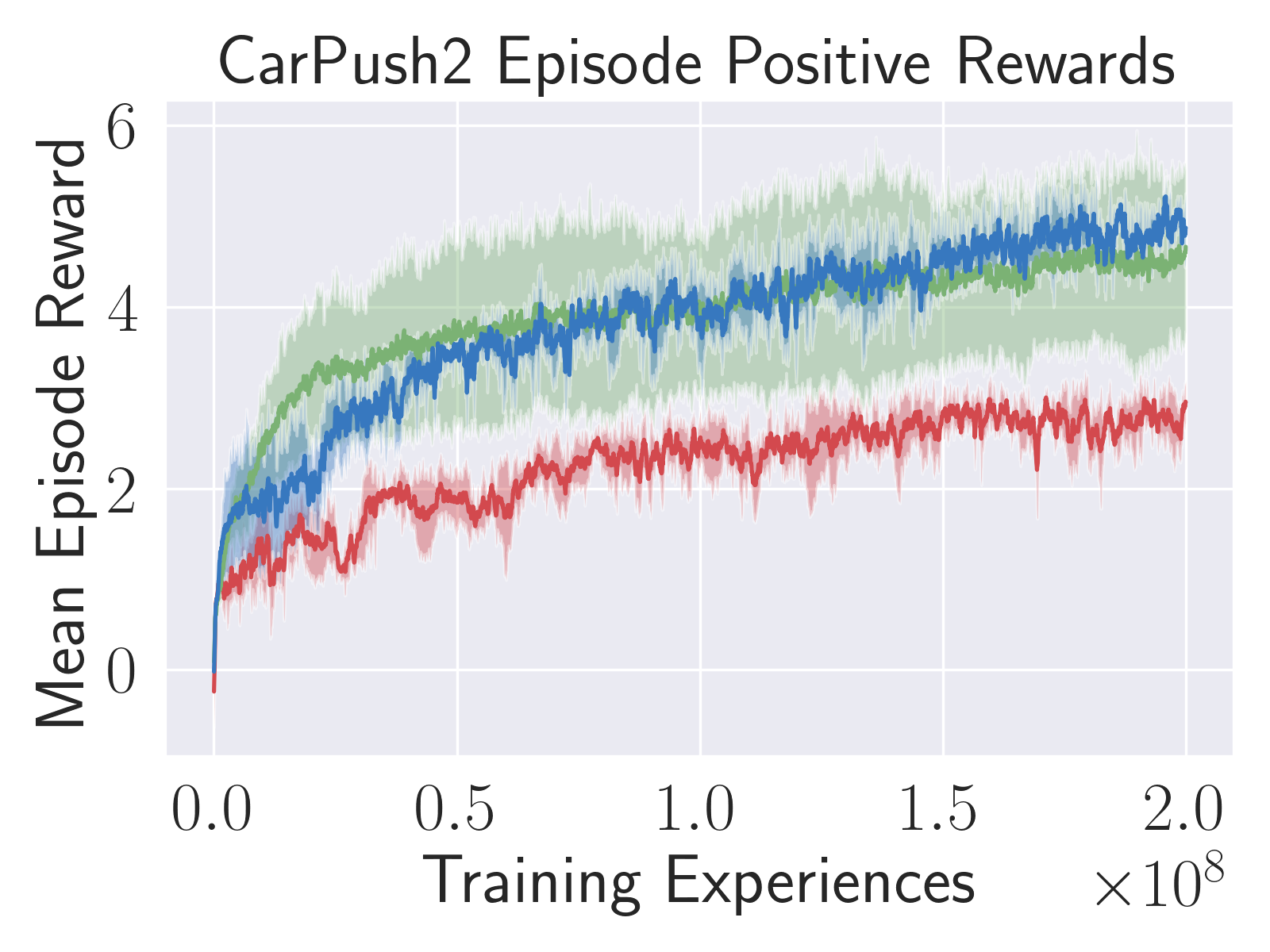}
\includegraphics[width=0.32\textwidth]{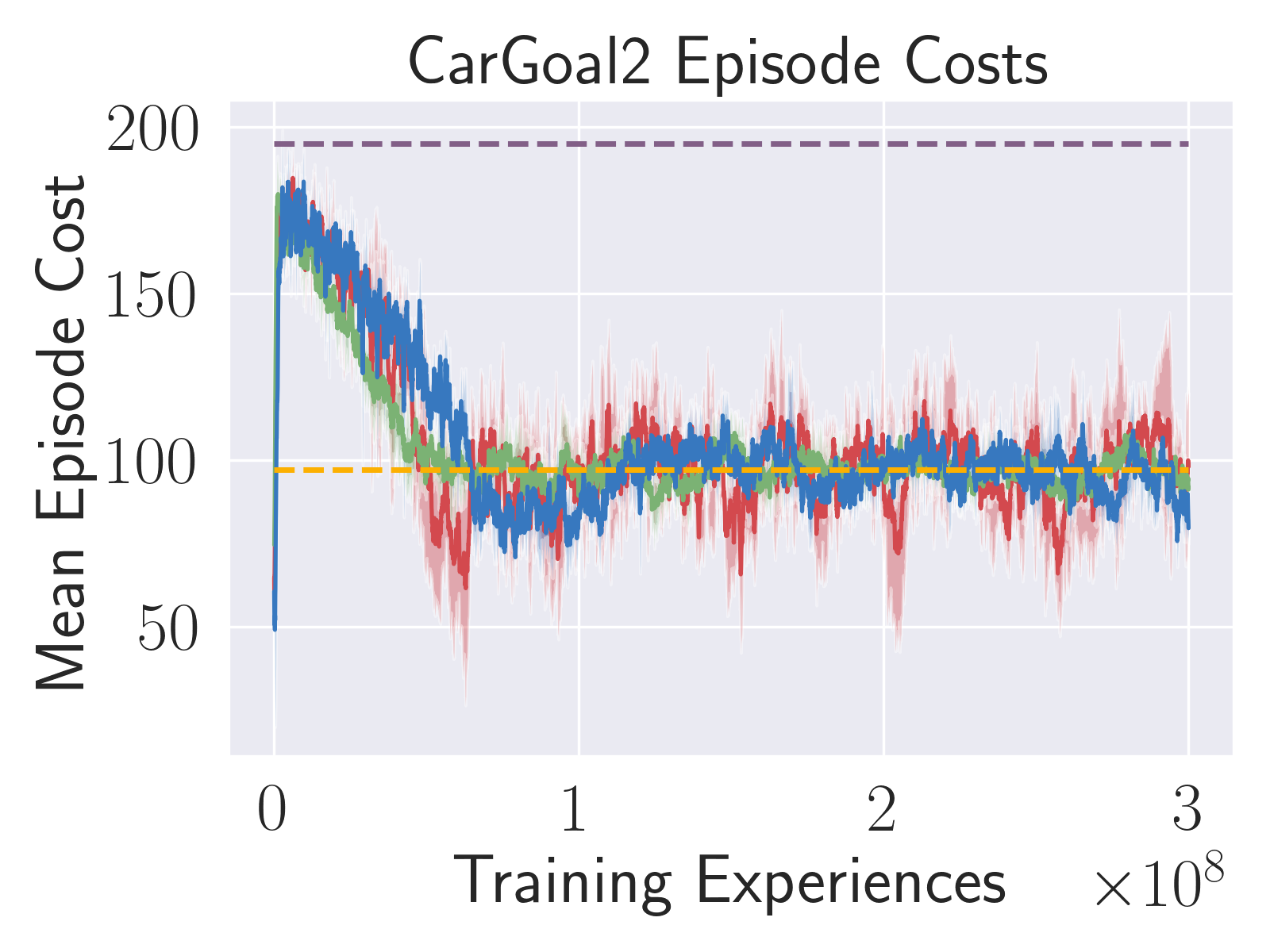}
\includegraphics[width=0.32\textwidth]{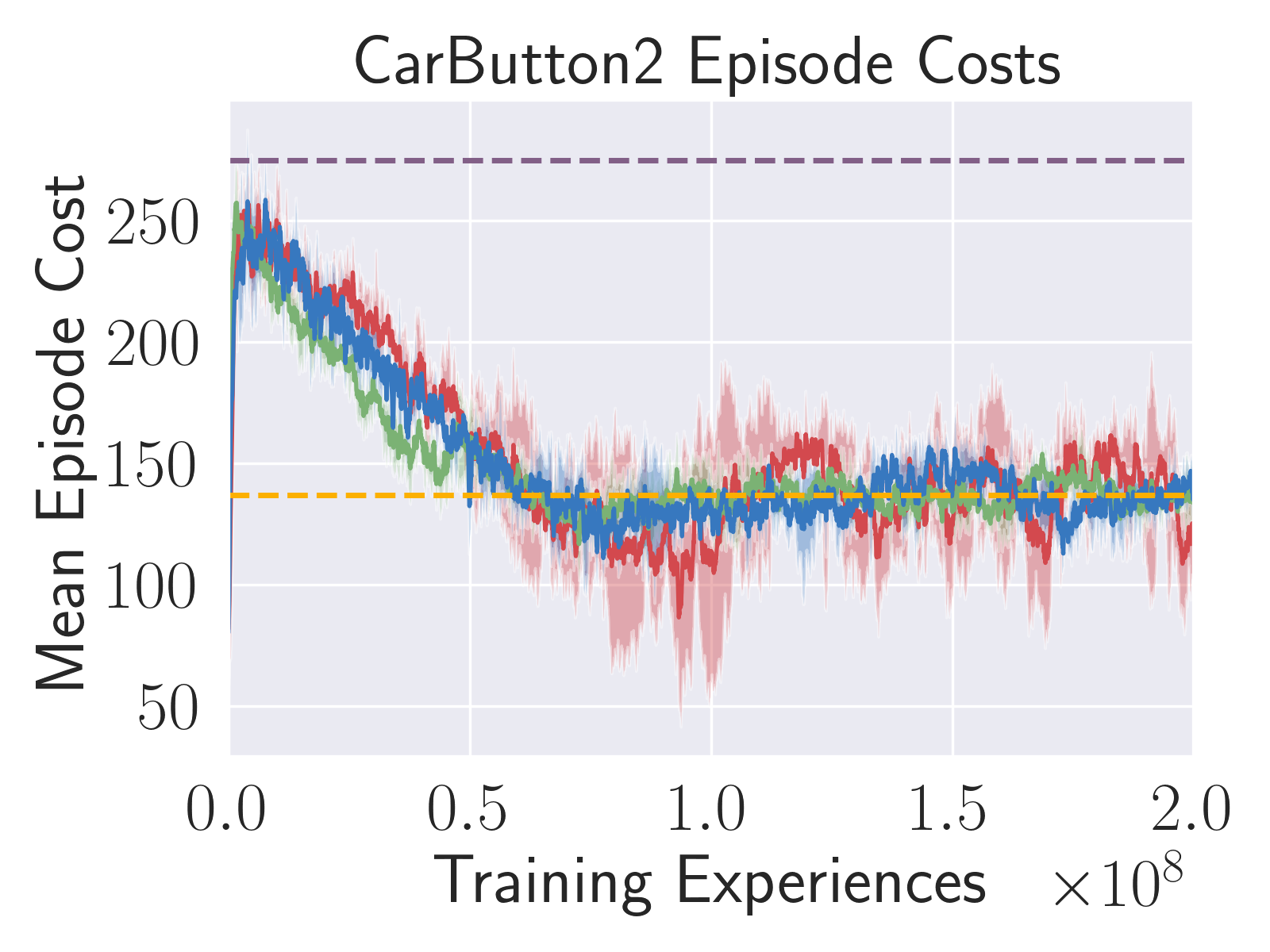}
\includegraphics[width=0.32\textwidth]{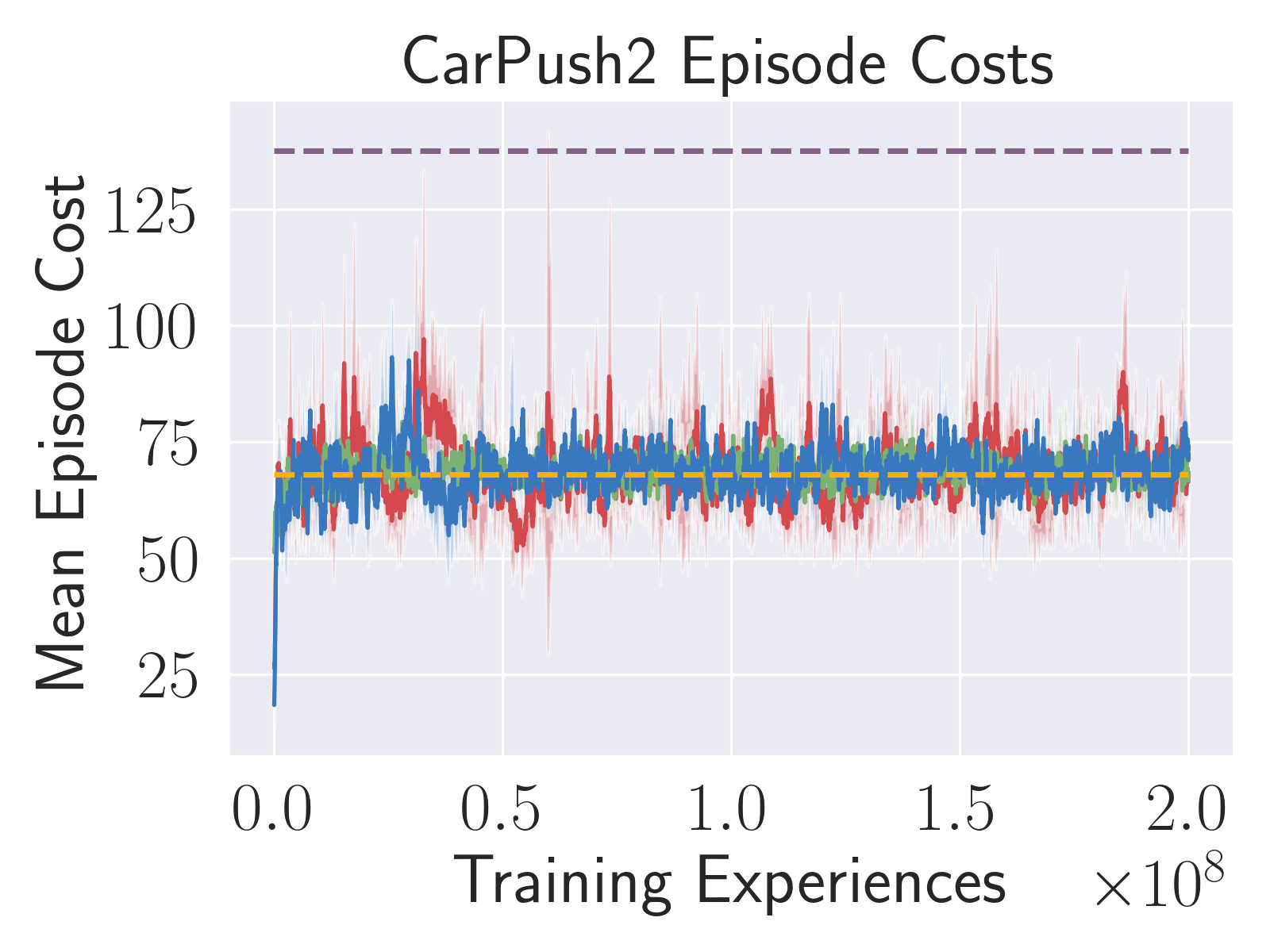}
\includegraphics[width=0.9\linewidth]{figures/legend_6.png}
\caption{Comparison of RCPO constrained policy optimization driven by PPO, TRPO, and COPG for Car robot environments. RCPO COPG obtains positive rewards that are higher than those of PPO and comparable to those of TRPO, while converging to the target cost level as quickly as the others.}
\end{figure}

\begin{figure}[H]
\centering
\includegraphics[width=0.32\textwidth]{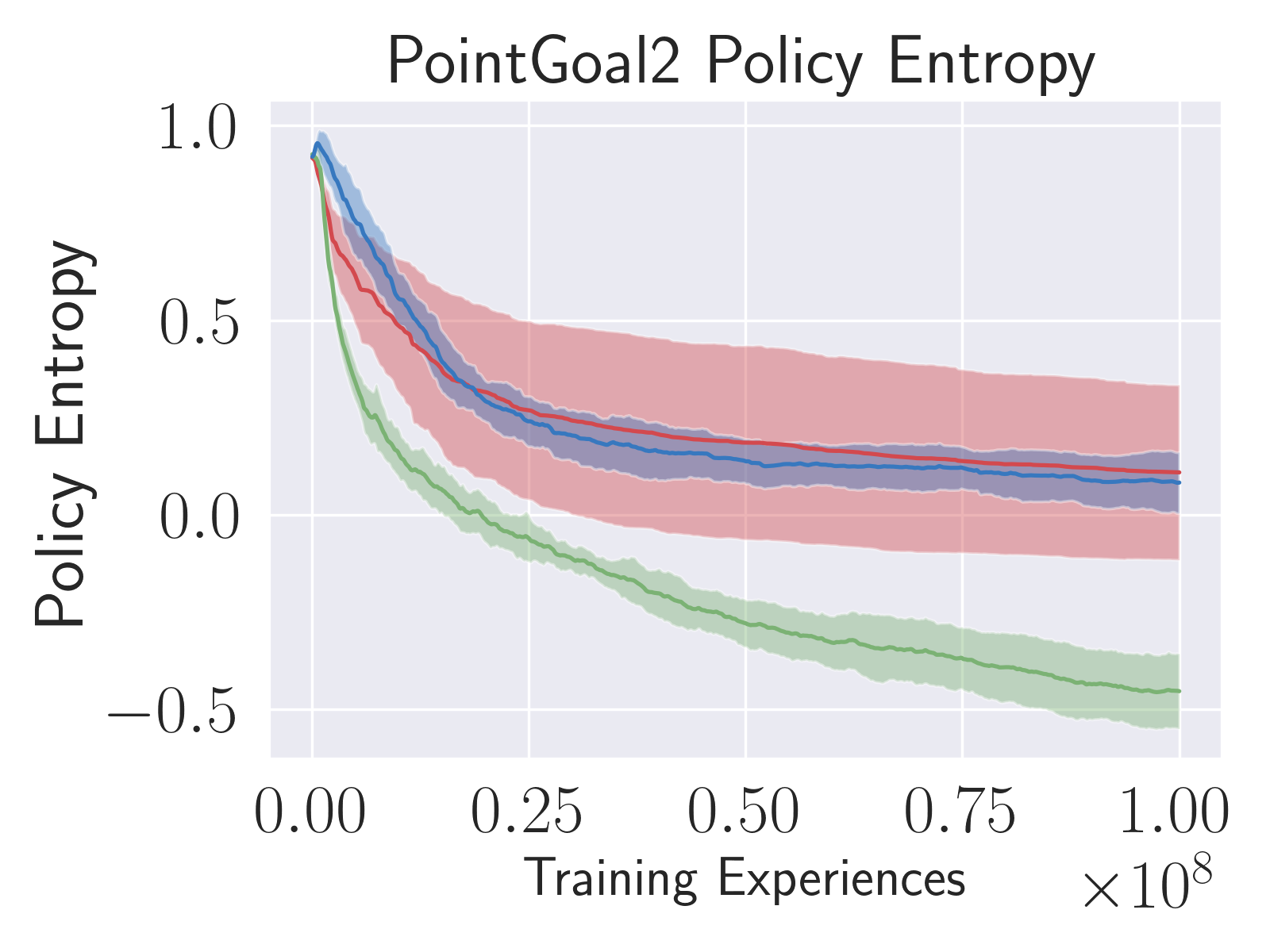}
\includegraphics[width=0.32\textwidth]{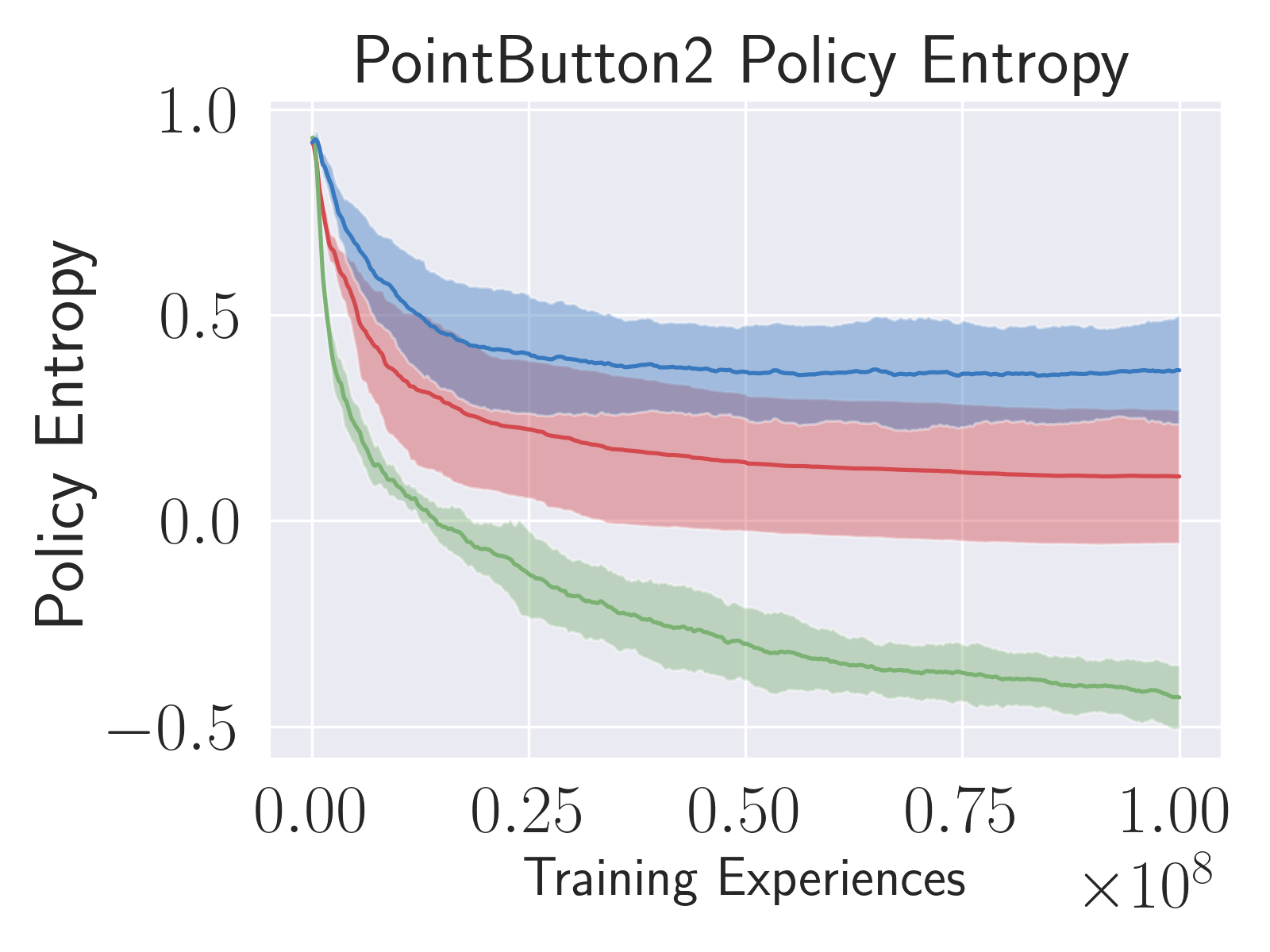}
\includegraphics[width=0.32\textwidth]{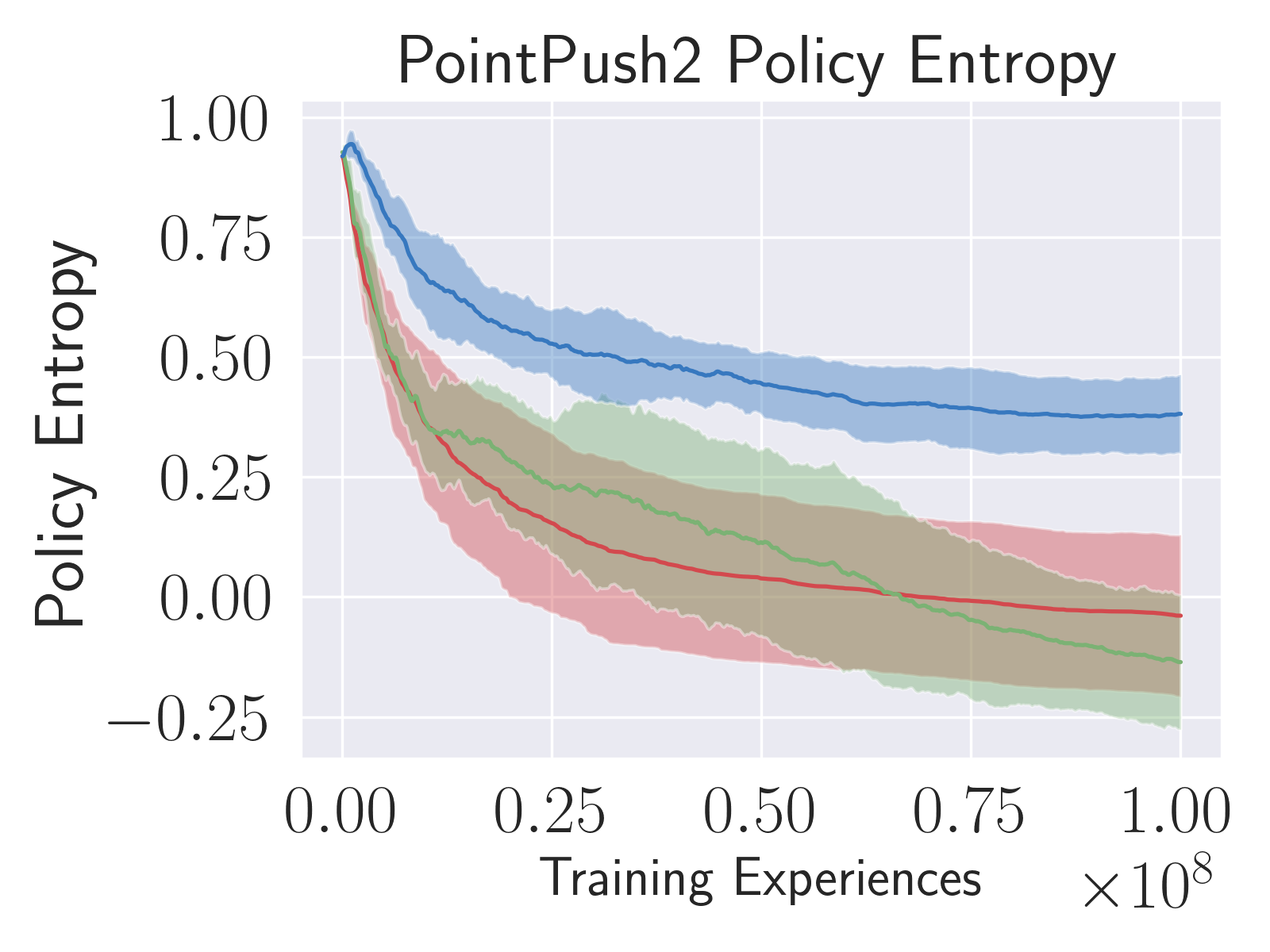}
\includegraphics[width=0.32\textwidth]{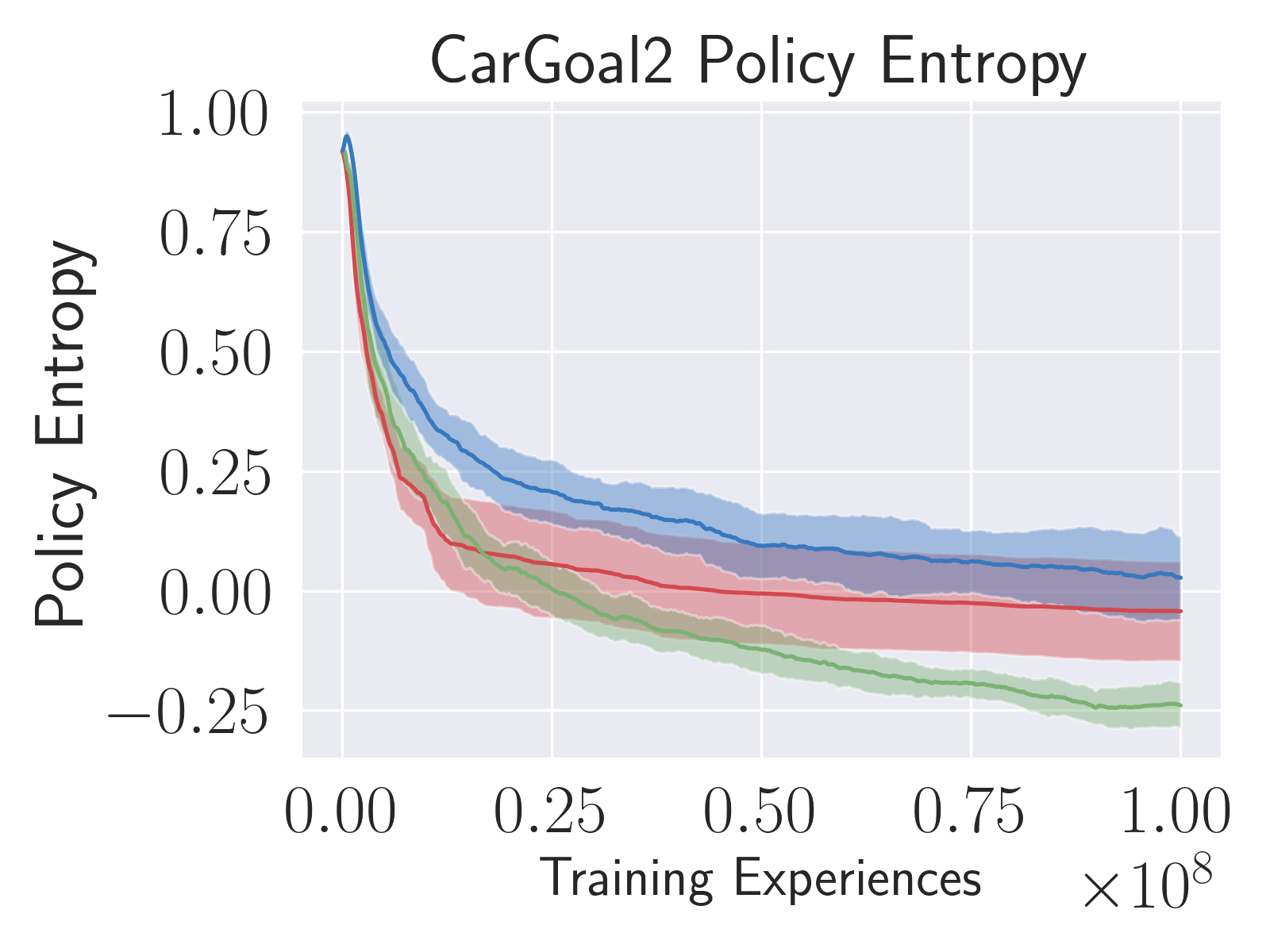}
\includegraphics[width=0.32\textwidth]{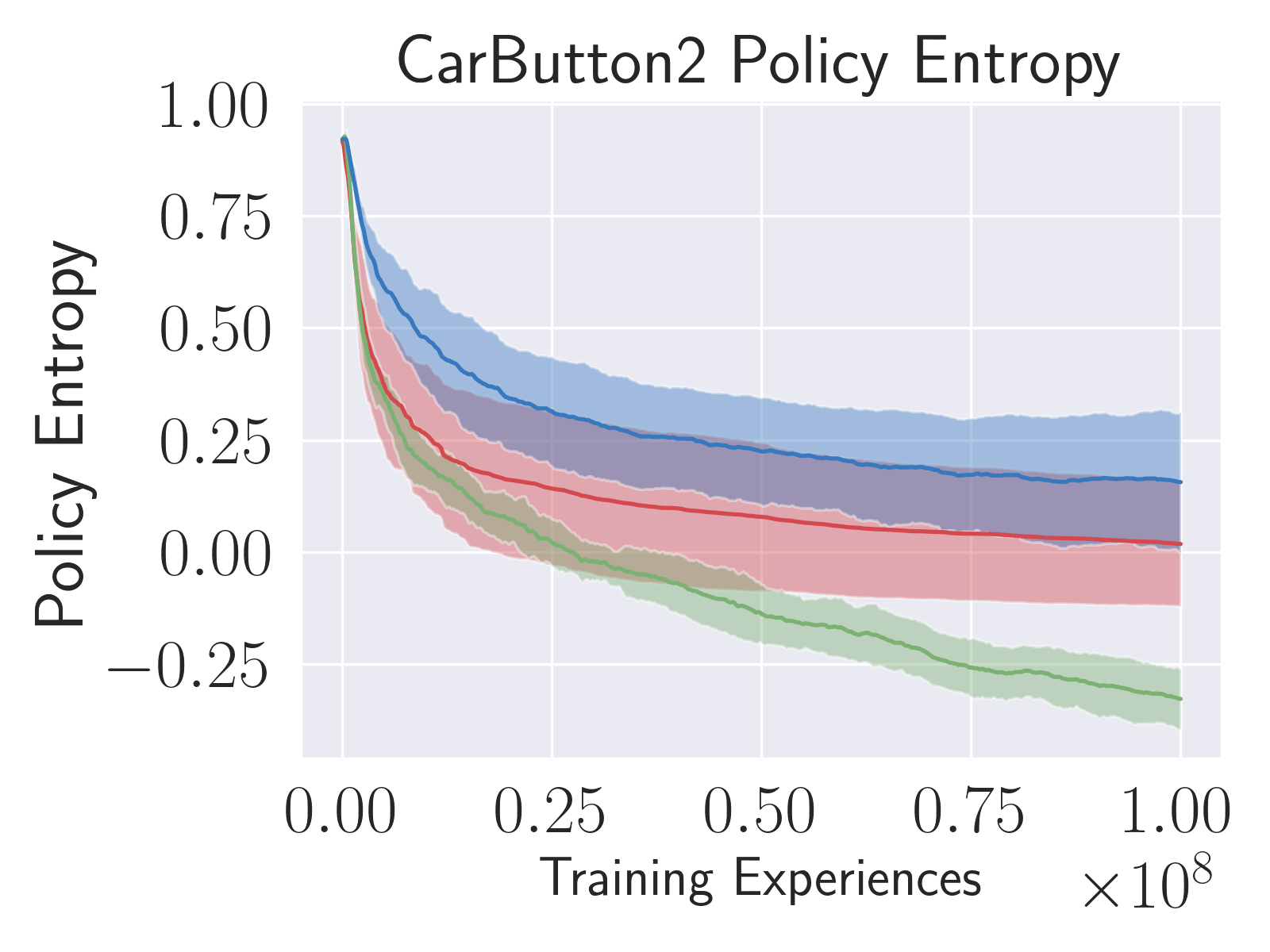}
\includegraphics[width=0.32\textwidth]{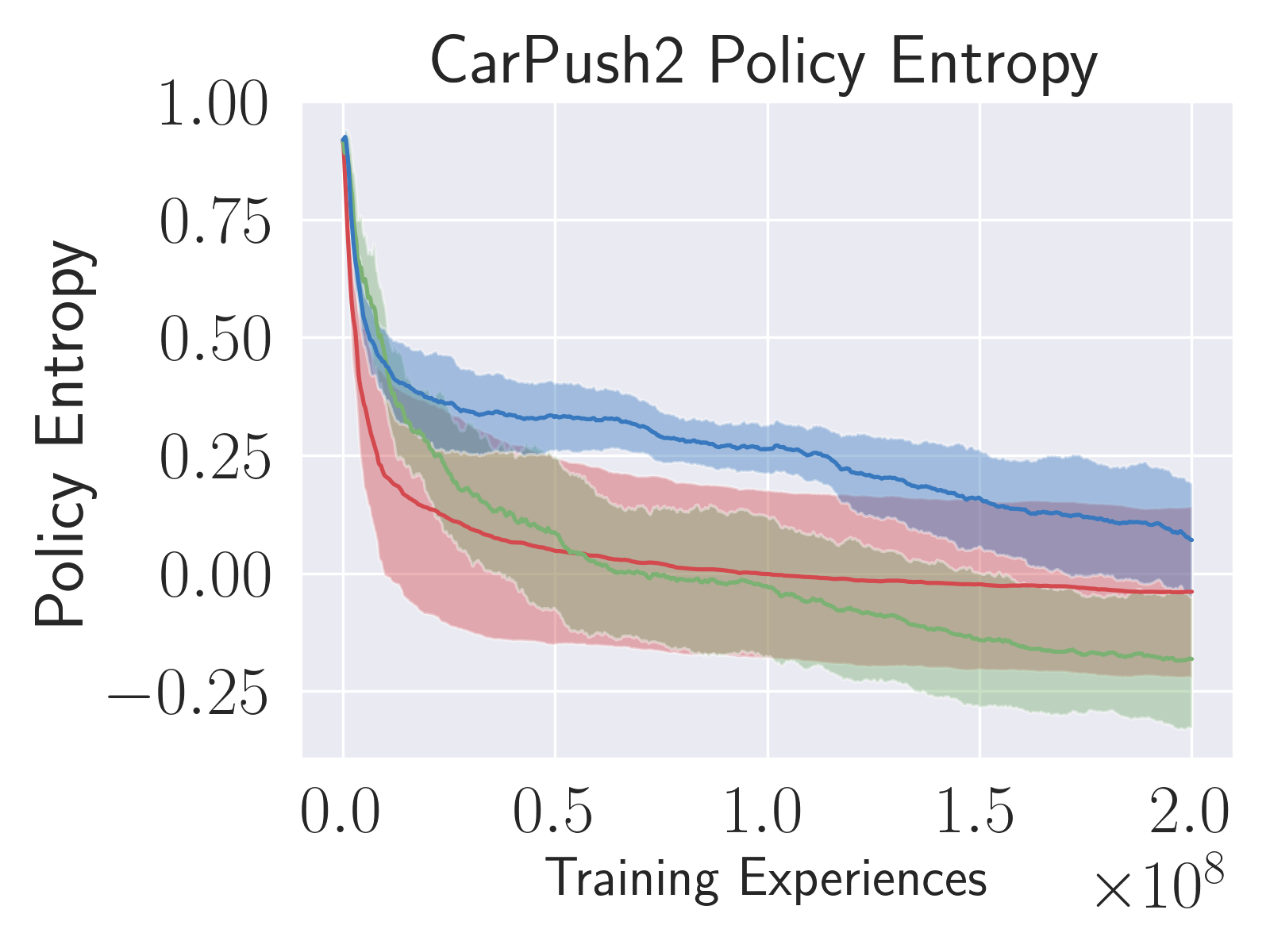}
\includegraphics[width=0.9\linewidth]{figures/legend_5.png}
\caption{Policy entropies for PPO, TRPO, and COPG in unconstrained learning of Safety Gym environments.  COPG consistently maintains higher entropy than PPO, aiding exploration.}
\end{figure}

\begin{figure}[H]
\centering
\includegraphics[width=0.32\textwidth]{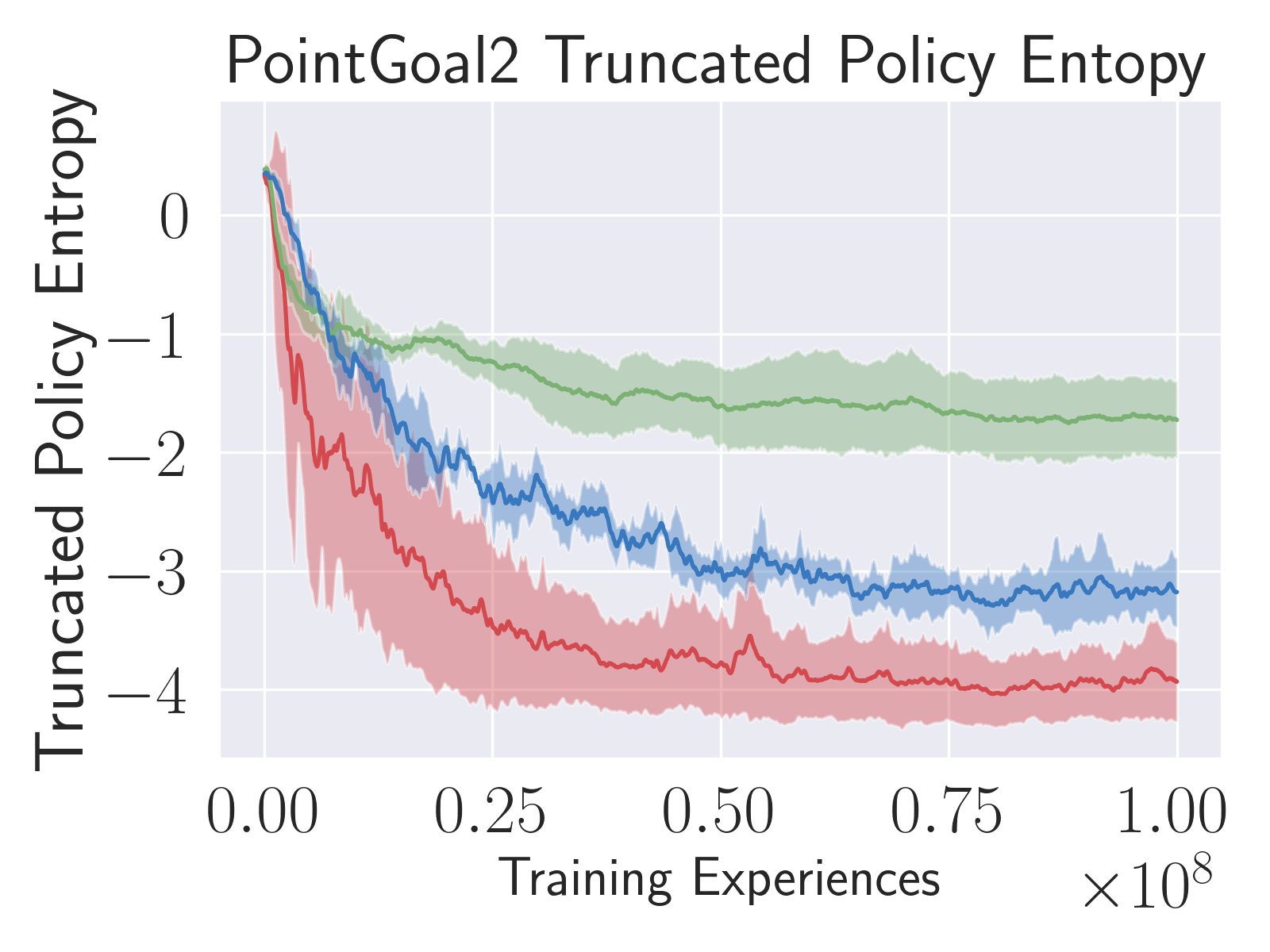}
\includegraphics[width=0.32\textwidth]{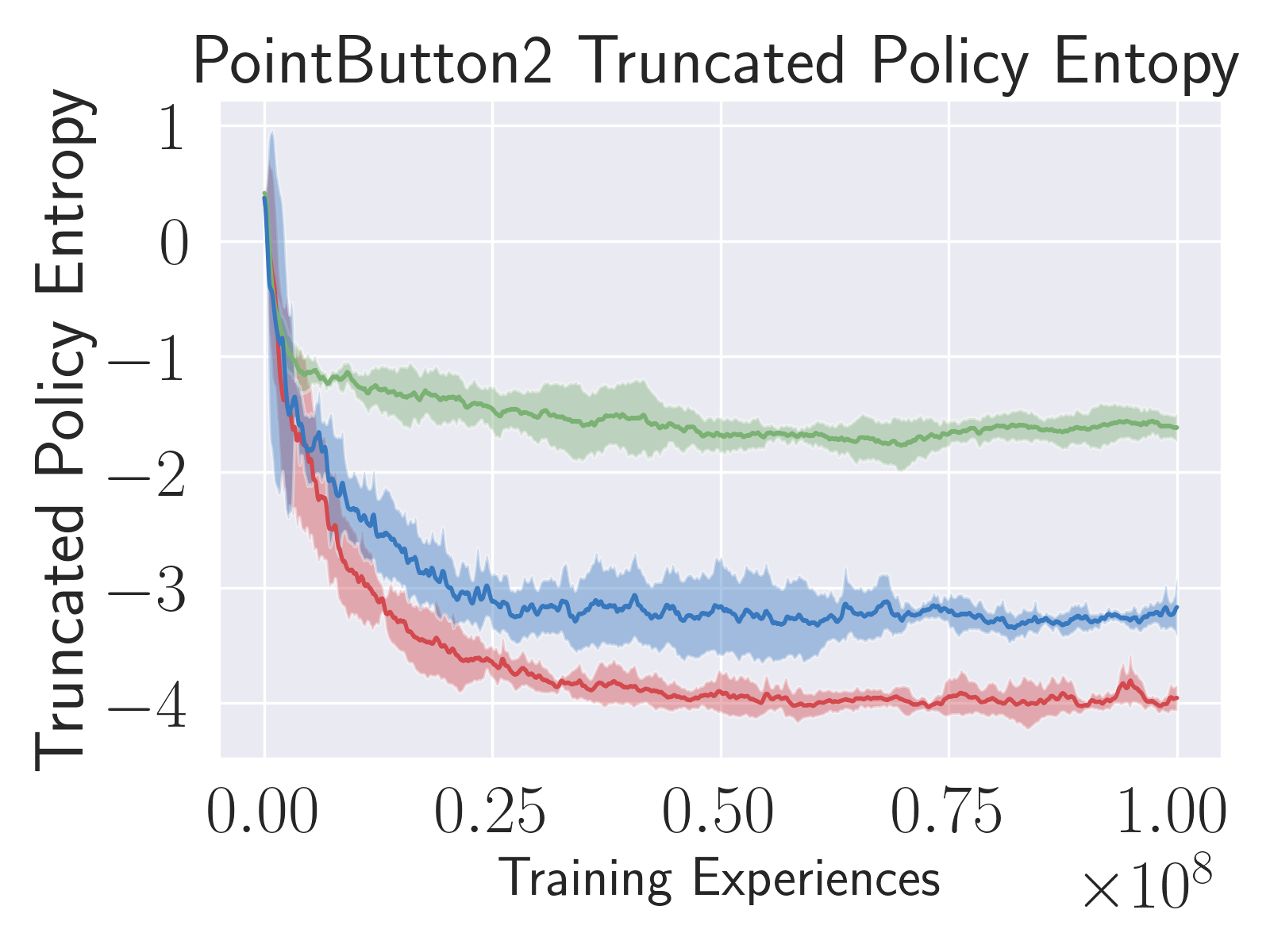}
\includegraphics[width=0.32\textwidth]{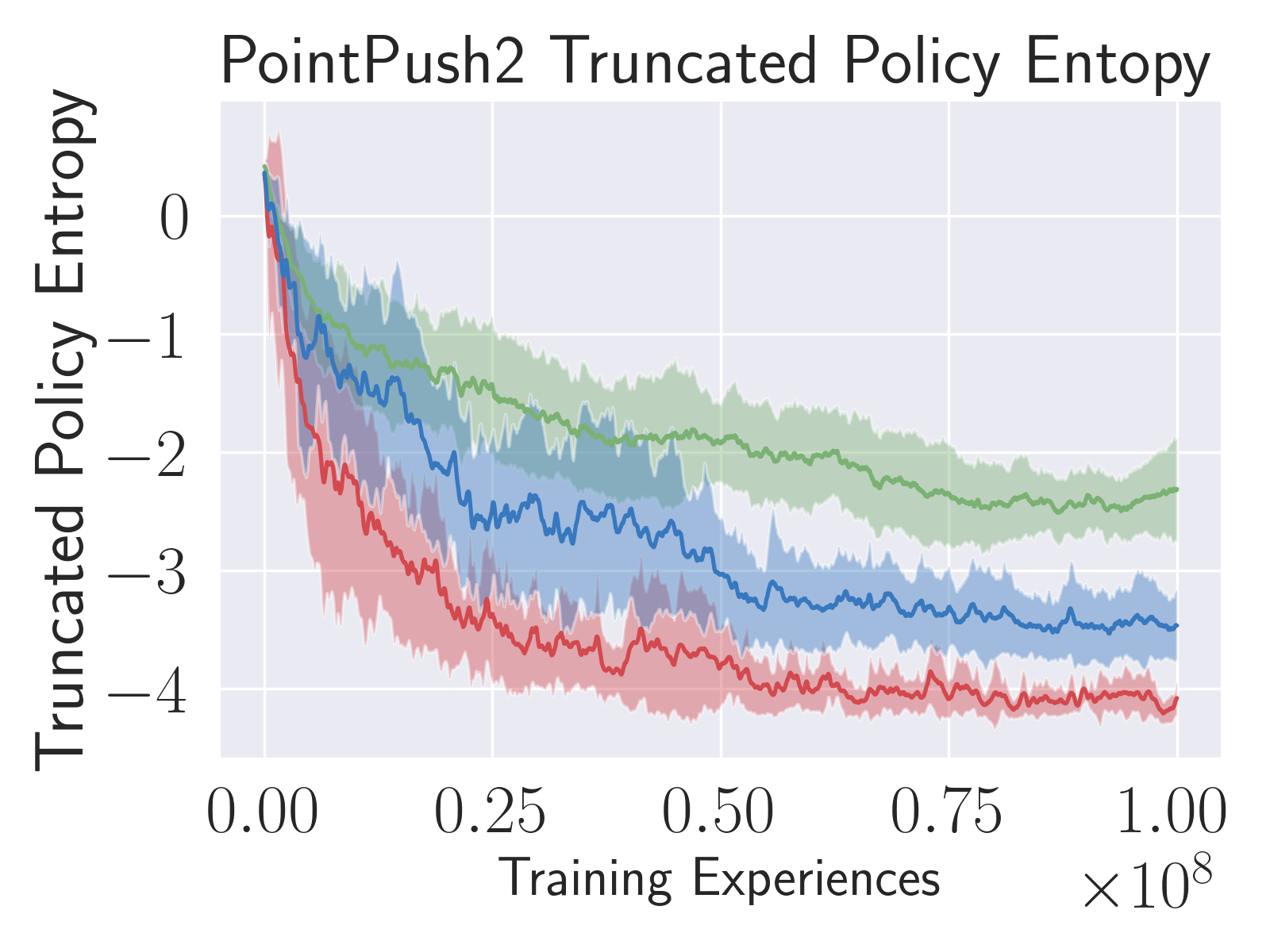}
\includegraphics[width=0.32\textwidth]{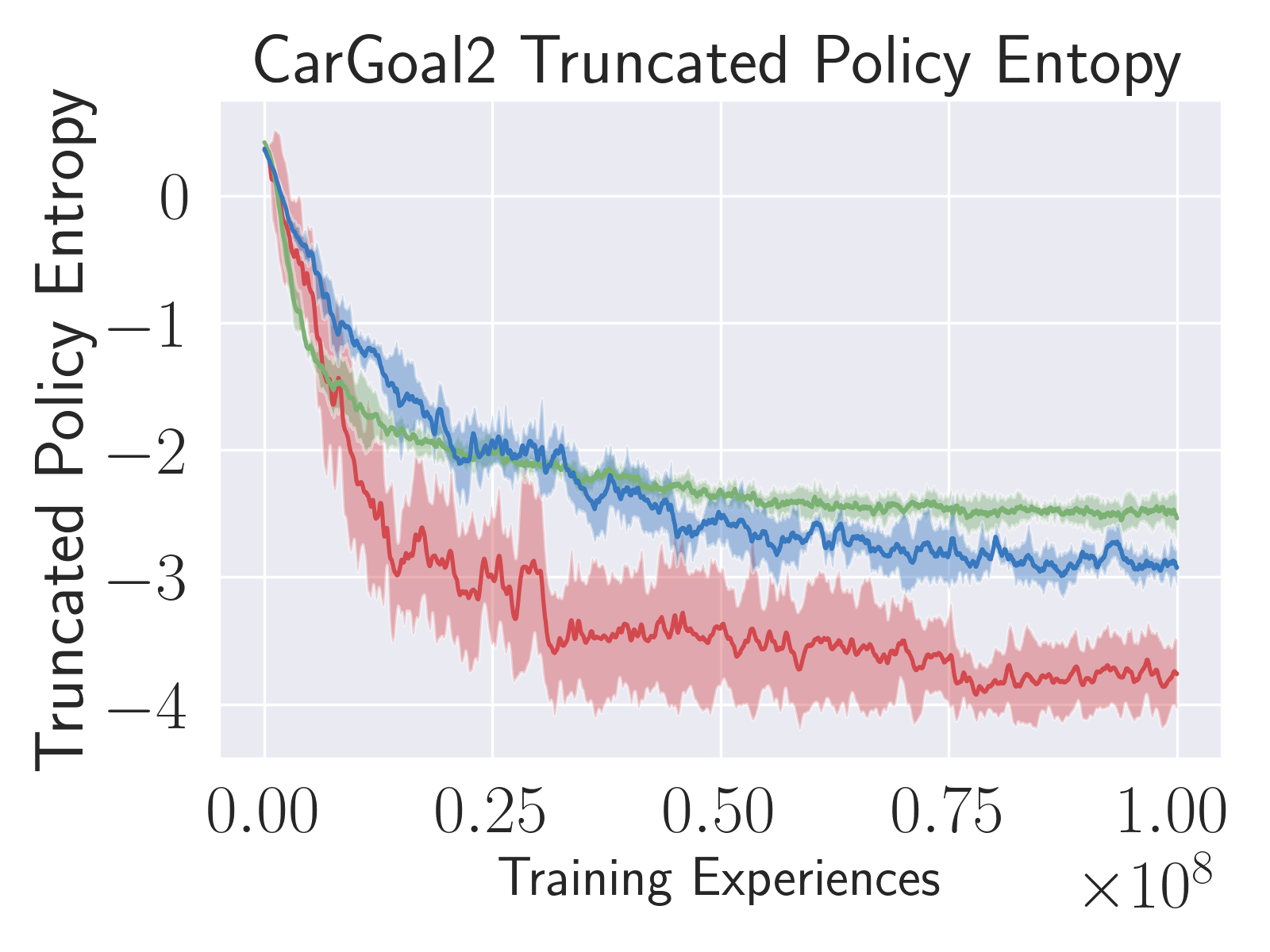}
\includegraphics[width=0.32\textwidth]{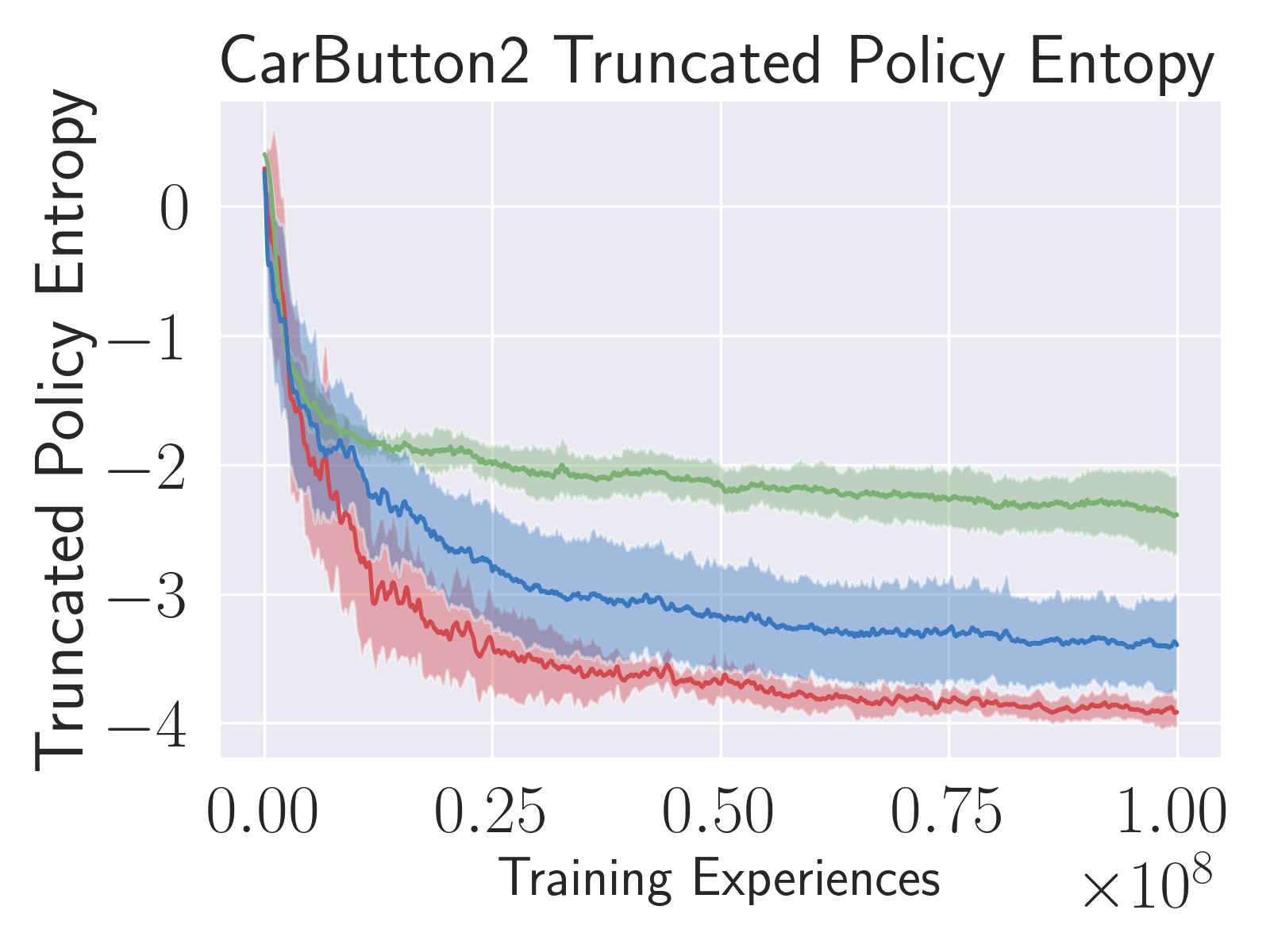}
\includegraphics[width=0.32\textwidth]{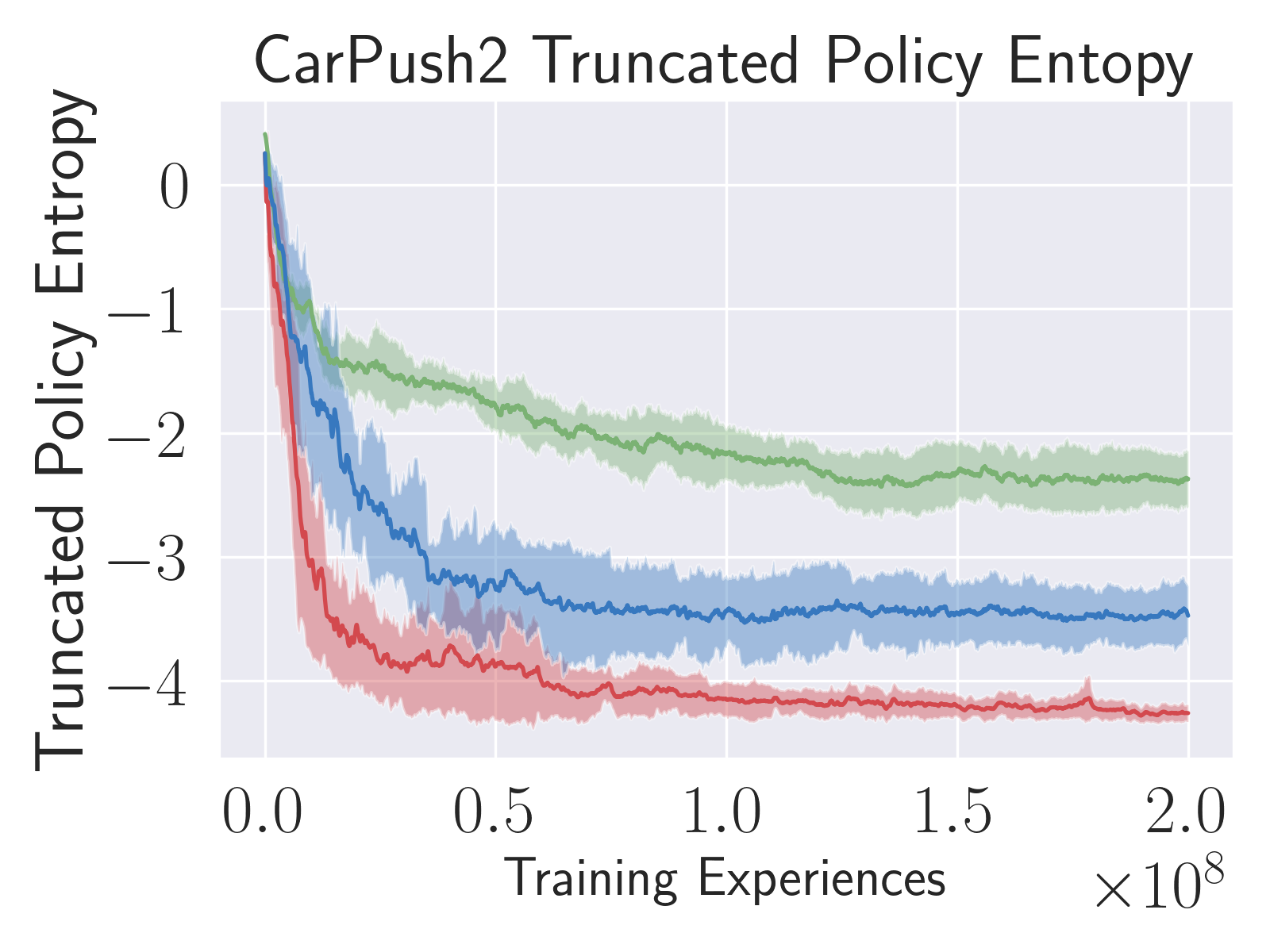}
\includegraphics[width=0.9\linewidth]{figures/legend_5.png}
\caption{Policy entropies for PPO, TRPO, and COPG in unconstrained learning of Safety Gym environments, taking into account control bounds.  COPG consistently maintains higher entropy than PPO, aiding exploration.}
\end{figure}

\end{document}